\documentclass[10pt,twocolumn,letterpaper]{article}

\usepackage{wacv}              

\definecolor{wacvblue}{rgb}{0.21,0.49,0.74}
\usepackage[pagebackref,breaklinks,colorlinks,allcolors=wacvblue]{hyperref}
\usepackage{graphicx}
\usepackage{booktabs}
\usepackage[accsupp]{axessibility}

\usepackage{url}
\usepackage[utf8]{inputenc}
\usepackage[T1]{fontenc}   
\usepackage{xspace}       
\usepackage{amsfonts}     
\usepackage{nicefrac}     
\usepackage{microtype}    
\usepackage{xcolor}         

\usepackage{multirow}
\usepackage{colortbl}
\usepackage{enumitem}
\usepackage{wrapfig}

\usepackage{rotating}
\usepackage{float}          
\usepackage{tabularx}       
\usepackage{makecell}       
\usepackage{caption}        

\usepackage{amsmath}
\usepackage{subcaption}

\newcommand{\ours}{{\textit{SenseShift6D}}\xspace}

\def\wacvPaperID{1263} 
\def\confName{WACV}
\def\confYear{2027}

\title{SenseShift6D: Multimodal RGB-D Benchmarking for Robust 6D Pose Estimation across Environment and Sensor Variations} 

\author{
Yegyu Han \quad
Taegyoon Yoon \quad
Dayeon Woo \quad
Sojeong Kim \quad
Hyung-Sin Kim%
\setcounter{footnote}{1}%
\thanks{Corresponding author}\\
Seoul National University\\
{\tt\small \{yegyuhan, taegyoun88, wuvele22, kvia2230, hyungkim\}@snu.ac.kr}
}

\begin{document}
\maketitle
\begin{abstract}
  Recent advances on 6D object pose estimation have achieved high performance on representative benchmarks such as LM-O, YCB-V, and T-Less.
  These datasets capture natural-occurring variations in illumination and camera setup but do not systematically explore the effects of controllable sensor configurations, such as exposure, gain, and depth-sensor modes.
  To bridge this gap, we introduce \ours, the first RGB-D object-pose benchmark that physically sweeps 13 RGB exposures, 9 RGB gains, auto-exposure, 4 depth-capture modes, and 5 illumination levels. For six common household objects, we acquire 198.8k RGB and 20.0k depth images (i.e., 795.4k RGB-D scenes), providing 1,380 unique sensor-lighting permutations per object pose.
Experiments with state‑of‑the‑art pretrained, generalizable pose estimators reveal substantial performance variation across lighting and sensor settings, despite their large‑scale pretraining. Strikingly, even instance‑level estimators---trained and tested on identical objects and backgrounds---exhibit pronounced sensitivity to environmental and sensor shifts. These findings establish sensor‑ and environment‑aware robustness as an underexplored yet essential dimension of 6D pose estimation.
To illustrate the opportunity enabled by this benchmark, we evaluate test‑time sensor selection without retraining. An idealized controller yields remarkable gains of up to +16.2 pp for generalizable models, whereas a practical consistency-based proxy provides only limited improvements, highlighting substantial headroom and the need for future research on reliable sensor‑aware adaptation. \ours is available at {\footnotesize\url{huggingface.co/datasets/Yegyu/SenseShift6D}}.
  
\end{abstract}    
\vspace{-2ex}
\section{Introduction}

Estimating the 6D pose of objects is a critical vision task, enabling fine-grained interactions in mixed-reality headsets, autonomous-vehicle manipulation, and embodied AI agents navigating cluttered environments~\cite{mr1, mr2, mr3, ad1, ad2, emAI1}.
The field has made substantial progress, with state-of-the-art methods achieving high accuracy on established benchmarks~\cite{LM, LMO, tless, hb, tudl}, even under challenging settings such as unseen objects and complex backgrounds.

Despite this progress, real-world deployments \textbf{rarely match} benchmark acquisition conditions. Illumination varies over time, camera exposure and gain are adjusted automatically or manually, and depth sensors operate under different capture modes on purpose.
These environmental and sensor dynamics introduce complex artifacts---nonlinear color shifts, saturation, and structured depth noise---that are difficult to reproduce via synthetic augmentations or correct reliably in post-processing~\cite{imagenet-es}.
However, most widely used benchmarks lack systematic variation in illumination and sensor configurations, leaving a key axis of robustness largely unmeasured: \textit{how sensitive are 6D pose estimators to changes in illumination and camera/depth‑sensor settings, even when the scene itself is simple?} While robustness to these covariate shifts have been recently studied in image classification tasks~\cite{imagenet-es,lens,kim2026imagenet,choi2026emerging}, the 6D pose community lacks a benchmark that systematically sweeps multimodal sensor parameters while providing accurate 6D ground truth.

 \begin{figure*}[t]
     \centering
     \includegraphics[width=\textwidth]{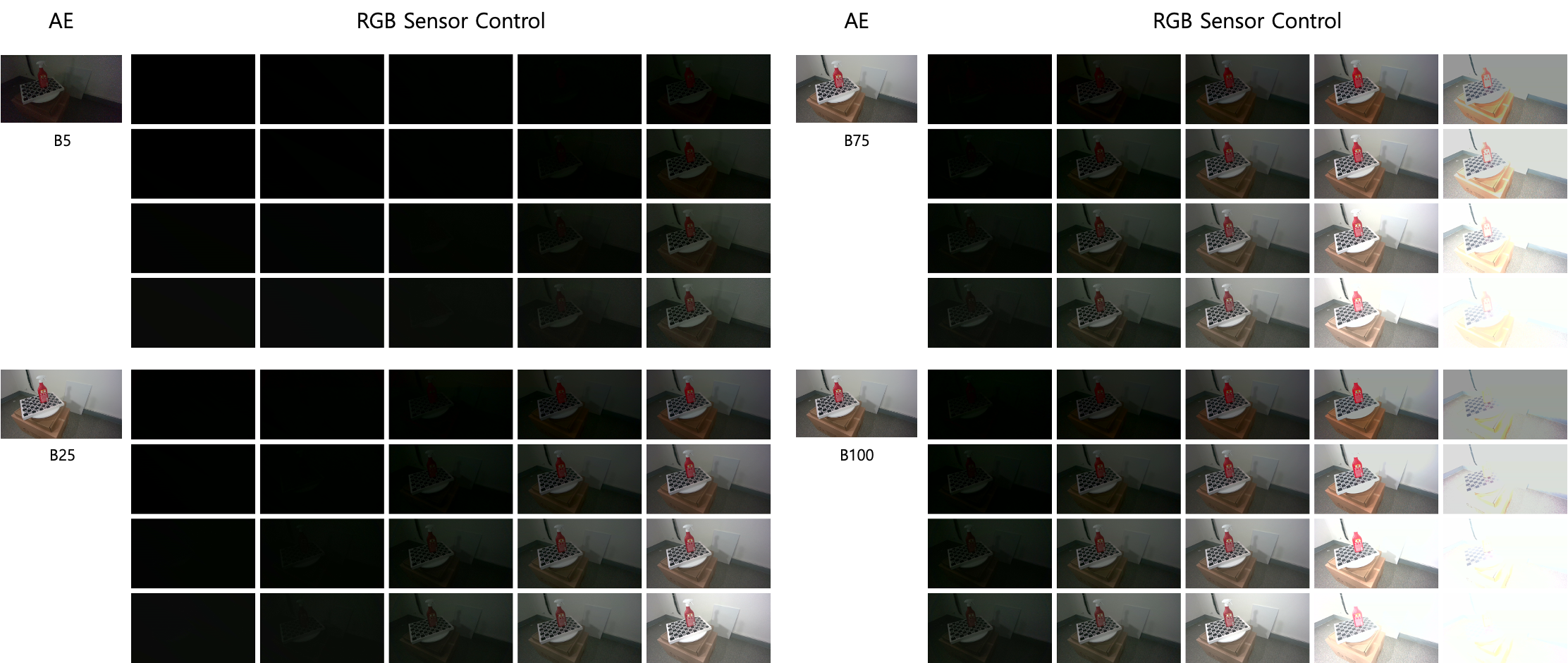}
     \vspace{-4.5ex}
     \caption{\textbf{RGB sample images under auto exposure and under all combinations of exposure and gain settings for brightness 5\%, 25\%, 75\% and 100\%.} Rows indicate gain levels and columns indicate exposure levels.}
     \vspace{-1ex}
     \label{fig:dataset-RGB-overview}
 \end{figure*}
 
  \begin{figure*}[t]
     \centering
     \includegraphics[width=\textwidth]{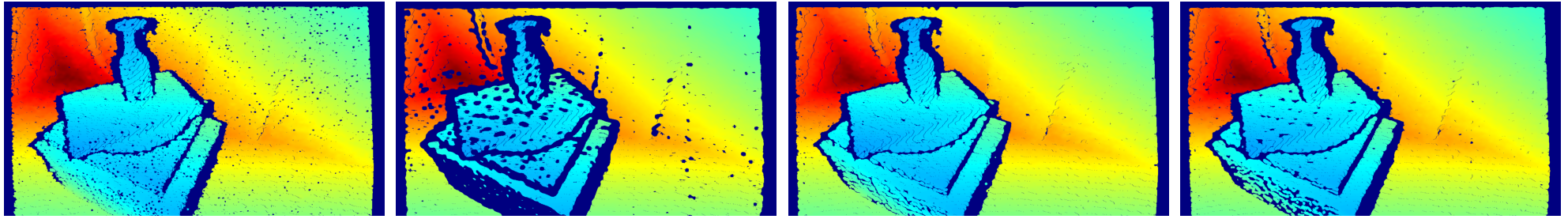}
     \vspace{-4.5ex}
     \caption{\textbf{Depth sample images under four depth capture modes:} default, high accuracy, high density, and medium density.}
     \vspace{-3ex}
     \label{fig:dataset-Depth-overview}
 \end{figure*}

To bridge this gap, we introduce \ours, a novel RGB-D benchmark for 6D pose estimation that \textbf{physically sweeps key sensor and lighting parameters}.
Using an Intel RealSense RGB-D camera in a controlled darkroom setup, we systematically vary 13 exposure levels, 9 gain settings, 4 depth-capture modes, and 5 illumination conditions across six representative household objects.
This protocol produces 198.8k RGB and 20.0k depth frames, which can provide 1,380 unique sensor-lighting permutations per object pose (\cref{fig:dataset-RGB-overview,fig:dataset-Depth-overview}), and 795.4k paired RGB-D scenes. To our knowledge, it provides \textbf{the largest variation} of sensor-lighting permutation in a 6D pose estimation dataset.
We further define standardized splits: \textbf{Train-Def}, captured under default camera and illumination settings, and \textbf{Train-Var}, covering a comprehensive multimodal parameter grid. Correspondingly, we construct two test splits (\textbf{Test-Def} and \textbf{Test-Var}) without overlap with the training configurations. 
As all variations are physically realized, the captured images preserve authentic sensor artifacts, including nonlinear color shifts, photon-shot noise, and structured depth errors. 

Using \ours, we provide the first systematic characterization of this environmental- and sensor‑shift axis across both (i) pretrained, generalizable pose estimators and (ii) instance‑level models trained on the same objects. 
We find that performance can change substantially across illumination and sensor settings---even for single‑object scenes with clean backgrounds---indicating that sensor/environment shifts constitute an \textbf{underexplored challenge} not covered by existing benchmarks or large‑scale pretraining alone. This establishes sensor‑aware evaluation as a necessary complement to the community's current emphasis on clutter, occlusion, and category generalization.
\ours also enables research on \textbf{sensor‑aware inference}~\cite{adaptivesensing}, including parameter selection at test time. In this work, we include simple sensor‑selection baselines and oracle upper bounds as diagnostic tools. Oracle selection reveals substantial headroom: single-modal control is \textit{already} effective (e.g., +9.1 percentage points (pp) on pretrained models), and \textbf{multimodal control} yields further improvements (e.g., +16.2 pp on the same models).
However, a realistic consistency‑proxy yields limited gains. This clear gap underscores the need for future research on reliable sensor-aware adaptation.

This paper offers three main contributions:
\begin{itemize}[leftmargin=*,noitemsep,topsep=0pt]
    \item \textbf{\ours benchmark:} We present the first 6D object-pose benchmark that systematically and physically varies multimodal sensor parameters and illumination, with accurate 6D ground truth and standardized train/test splits.
     
    \item \textbf{A new robustness axis for 6D pose:} 
    We reveal that illumination and sensor settings constitute an orthogonal axis of generalization. Both pretrained generalizable models and instance-level models degrade markedly under these shifts, even in low-complexity scenes.

    \item \textbf{Sensor-aware evaluation protocol and baselines:} 
    We provide a structured evaluation suite and sensor-selection diagnostics (upper bounds and realistic proxies) to quantify robustness and headroom, enabling principled future work on sensor-aware 6D pose estimation. 
\end{itemize}

With a structured testbed for sensor-aware object pose estimation, \ours lays the groundwork for more robust, adaptive perception in dynamic real-world environments.
\section{Related Work}
\begin{table}[t]
\centering
\caption{\textbf{Comparison of 6D pose estimation datasets.}}
\vspace{-2ex}
\resizebox{\linewidth}{!}{  
\begin{tabular}{lccccc}
\toprule
\multirow{2}{*}{\raisebox{-0.6ex}{\textbf{Dataset}}} & \multicolumn{5}{c}{\textbf{Environmental and sensor variations}} \\
\cmidrule(lr){2-6}
 & \textbf{Illumination} & \textbf{Exposure} & \textbf{Gain} & \textbf{Depth} & \textbf{Total Var.}  \\
\midrule
LM~\cite{LM}        & $\times$ & $\times$ & $\times$ & $\times$ & $\times$ \\
LM-O~\cite{LMO}     & $\times$ & $\times$ & $\times$ & $\times$ & $\times$  \\
T-LESS~\cite{tless} & $\times$ & $\times$ & $\times$ & $\times$ & $\times$ \\
YCB-V~\cite{PoseCNN} & $\times$ & $\times$ & $\times$ & $\times$ & $\times$  \\
TUD-L~\cite{tudl}   & \textbf{8 options} & $\times$ & $\times$ & $\times$ & 8 \\
HB~\cite{hb}        & 2 options & $\times$ & $\times$ & $\times$ & 2 \\
HOPE~\cite{hope}    & 5 options & $\times$ & $\times$ & $\times$ & 5 \\
EgoXtreme~\cite{yoon2026egoxtreme}    & \textbf{8 options} & $\times$ & $\times$ & $\times$ & 8 \\
IPD~\cite{ipd}      & 3 options & 4 levels & $\times$ & $\times$ & 12 \\
\midrule
\rowcolor[HTML]{E3F2FD}
 \textbf{SenseShift6D (Ours)}  & 5 levels    & \textbf{13 levels}& \textbf{9 levels} & \textbf{4 modes}  &  \textbf{1,380} \\
\bottomrule
\end{tabular}}
\vspace{-2ex}
\label{tab:sensor_variation}
\end{table}

\subsection{6D Pose Estimation Benchmark}

\noindent\textbf{Classic datasets.} 
LineMOD (LM) \cite{LM} and its occlusion subset (LM-O) 
\cite{LMO} pioneered RGB-D pose estimation benchmarks in controlled tabletop setups.
Subsequent benchmarks such as T-LESS \cite{tless}, which provided a large collection of texture-less industrial objects, and YCB-V \cite{PoseCNN}, which offered cluttered scenes with temporal sequences, expanded the scope of evaluation.
While these benchmarks provide robust evaluation regarding variations in objects, all were captured under fixed illumination and camera settings.

\noindent\textbf{Initial exploration of environmental and sensor variation.} 
TUD-L \cite{tudl}, HomebrewedDB (HB) \cite{hb}, HOPE \cite{hope}, and EgoXtreme~\cite{yoon2026egoxtreme} varied illumination conditions but retained fixed capture configuration with auto-exposure enabled. 
The recent Industrial Parts Dataset (IPD)~\cite{ipd} extends 6D pose evaluation to camera and HDR co-evaluation, capturing four exposure levels under multiple lighting conditions. However, IPD focuses on texture-less industrial parts and does not explore variations in RGB gain or depth-sensor modes, thus limiting its suitability for adaptive-sensing research.

\noindent\textbf{Remaining gap and our contribution.} 
As in \cref{tab:sensor_variation}, no benchmark offers joint sweeps of RGB exposure, RGB gain, and multiple depth-capture modes with dense 6D ground truth. \ours addresses this gap by systematically varying these sensor parameters under diverse illumination conditions, enabling fine-grained analysis of 6D pose estimation robustness to realistic hardware-level changes.

\subsection{Adaptive Sensor Control Benchmark}
Prior studies~\cite{lidarHPO, camtuner} have explored test-time adaptation of camera parameters to improve downstream vision performance, but lacked dedicated real-world benchmarks.
Recently, ImageNet-ES~\cite{imagenet-es} and ImageNet-ES-Diverse~\cite{lens} captured real-world scenes using a \textit{physical} camera, systematically varying aperture, shutter speed, and ISO across multiple lighting conditions for image classification tasks. 
These benchmarks demonstrated that physical variation \textit{cannot be replicated by synthetic augmentations} and adaptive parameter selection at test time can significantly improve the performance of lightweight models, enabling accuracy competitive with larger foundation models without additional training.

However, existing adaptive sensing research remains limited to RGB classification~\cite{choi2026emerging,choiorganized,im2026measurement,suh2026resource}. No prior work has explored multimodal sensor control, where depth-sensor modes are jointly optimized with RGB settings, or the impact of sensor control on 6D pose estimation, which inherently relies on both photometric and geometric cues. 
Our \ours addresses this gap by providing the first testbed for evaluating multimodal per-frame sensor-control policies alongside conventional data-centric training baselines.

\begin{figure*}[t]
\centering

\begin{minipage}[t]{0.7\textwidth}
    \vspace{0ex}
    \centering
    \captionof{table}{\textbf{SenseShift6D Configurations for Data split.}
    Auto images for each split are processed from multiple auto-exposure shots.}
    \vspace{-2ex}
    \label{tab:capture-settings}
    \small
    \resizebox{\linewidth}{!}{
    \begin{tabular}{lccccccc}
    \toprule
    \textbf{Data split} &
    \textbf{Brightness (\%)} &
    \textbf{Sensor configuration} &
    \textbf{Exposure (\boldmath$\mu$s)} &
    \textbf{Gain} &
    \textbf{Depth-capture mode} &
    \textbf{Captured images} \\
    \midrule
    
    \makecell{Train-Def}
    & \makecell{50}
    & \makecell{Auto (1 RGB options,\\ 1 depth option)}
    & \makecell{Auto}
    & \makecell{Auto}
    & \makecell{Default}
    & \makecell{RGB: 750, Depth: 750\\1 RGB-D scene/pose} \\
    \midrule
    
    \makecell{Train-Var}
    & \makecell{5, 25, \\ 50, 75, 100}
    & \makecell{Auto + Manual \\ (41 RGB options,\\ 1 depth options)}
    & \makecell{Auto, 2, 4, 19, \\ 78, 312, 1250, \\ 5000, 10000}
    & \makecell{Auto, 0, 32, \\ 64, 96, 128}
    & \makecell{Default } 
    & \makecell{RGB: 153.7k, Depth: 3.7k\\ 205 RGB-D scenes/pose}  \\
    \midrule
    
    \makecell{Test-Def}
    & \makecell{50}
    & \makecell{Auto (1 RGB option,\\ 1 depth option)}
    & \makecell{Auto}
    & \makecell{Auto}
    & \makecell{Default}
    & \makecell{RGB: 251, Depth: 251\\1 RGB-D scenes/pose} \\
    \midrule
    
    \makecell{Test-Var}
    & \makecell{5, 25, \\ 50, 75, 100}
    & \makecell{Auto + Manual \\ (21 RGB options,\\ 4 Depth options)}
    & \makecell{Auto, 9, 39,\\ 156, 625, 2500}
    & \makecell{Auto, 16, 48, \\ 80, 112 }
    & \makecell{Default, \\ High Accuracy, \\ High Density, \\ Medium Density}
    & \makecell{RGB: 26.3k, Depth: 5.0k\\420 RGB-D scenes/pose}  \\
    \bottomrule
    \end{tabular}
    }
\end{minipage}
\hfill
\begin{minipage}[t]{0.27\textwidth}
    \vspace{5ex}
    \centering
    \includegraphics[width=\linewidth]{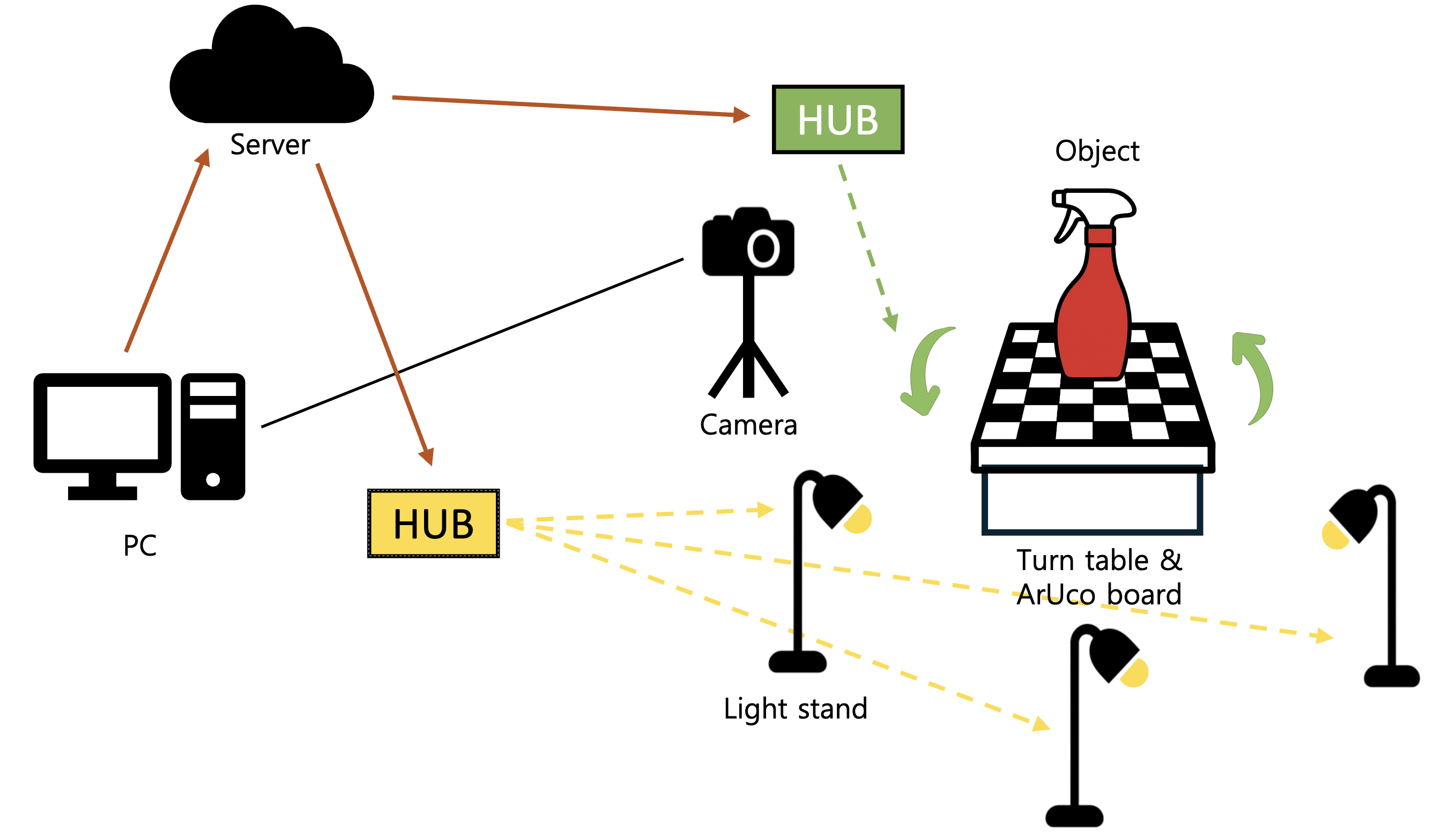}
    \captionof{figure}{\textbf{Testbed for data collection.}}
    \label{fig:testbed}
\end{minipage}
\vspace{-2ex}
\end{figure*}

\section{SenseShift6D}

\begin{figure}
  \centering
  \includegraphics[width=0.8\linewidth]{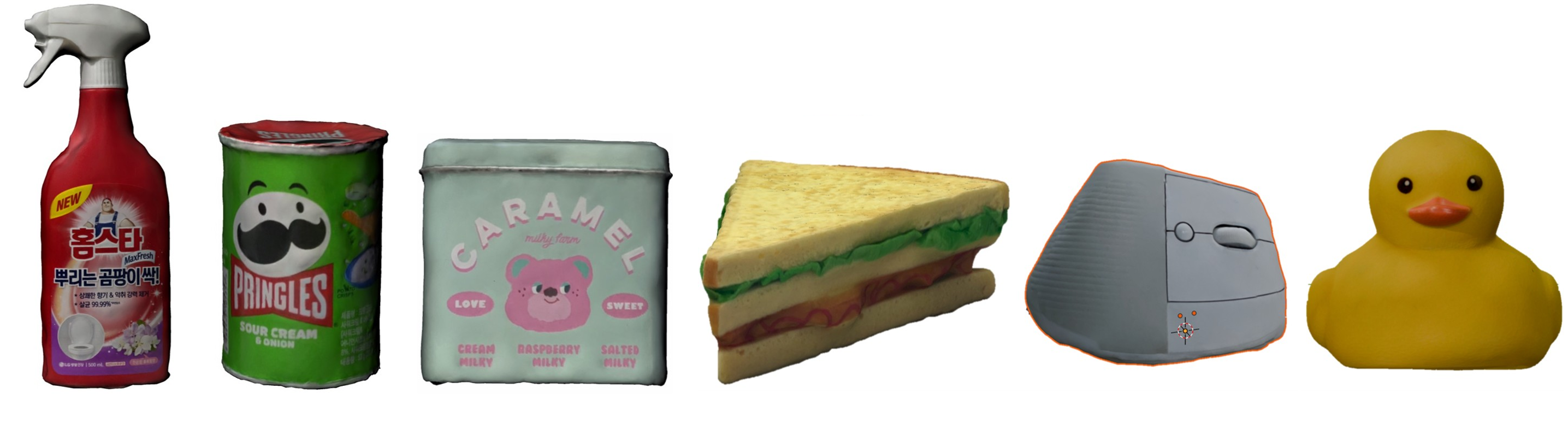}
  \vspace{-3ex}
     \caption{\textbf{Six household objects collected in SenseShift6D.}}
     \vspace{-3ex}
     \label{fig:object}
\end{figure}

We introduce \ours, the first RGB-D benchmark that \textit{physically} sweeps camera and lighting parameters, enabling rigorous evaluation of 6D object pose estimators under realistic environmental and sensor dynamics. 
Using an Intel RealSense D455 mounted in a darkroom, we systematically capture six household objects---spray, Pringles, tincase, sandwich, mouse, and duck---selected to reflect diverse combinations of RGB and geometric features (\cref{fig:object}), under auto-exposure, 13 RGB-exposure times, 9 RGB-gain settings, 4 depth-capture presets, and 5 illumination levels (details in \cref{tab:capture-settings}).
Our dataset contains 198.8k RGB and 20.0k depth frames, resulting in 1,380 unique sensor-lighting configurations per object pose (i.e., 5 brightness levels $\times$ 4 depth modes $\times$ (45 RGB for training + 24 RGB for testing)), which is the \textit{largest sensor-lighting configuration space} among the current benchmarks to our knowledge, and 795.4k RGB-D scene instances (i.e., 5 brightness level $\times$ 4 depth modes $\times$ (750 object poses $\times$ 45 RGB for training + 251 object poses $\times$ 24 RGB for testing)).

To facilitate controlled benchmarking, we organize the dataset into four clearly defined splits: 
\begin{itemize}[leftmargin=*,noitemsep,topsep=0pt]
    \item \textbf{Train-Def} contains only the default sensor and lighting configuration (1 RGB-D scene/pose)

    \item \textbf{Train-Var} expands Train-Def, covering (8 exposures $\times$ 5 gains + 1 auto-exposure) $\times$ 1 depth mode $\times$ 5 brightness levels (205 RGB-D scenes/pose). 

    \item \textbf{Test-Def} includes the default sensor and lighting configuration (1 RGB-D scene/pose)
    
    \item \textbf{Test-Var} expands Test-Def, featuring (5 exposures $\times$ 4 gains + 1 auto-exposure) $\times$ 4 depth-modes $\times$ 5 brightness levels (420 RGB-D scenes/pose). 
\end{itemize}

As all variations are physically captured rather than digitally simulated, the resulting images inherently retain authentic artifacts, such as non-linear color shifts, photon-shot noise, and depth-mode-specific errors, that synthetic augmentation methods cannot reproduce~\cite{imagenet-es}. 
\ours therefore offers a controlled testbed for analyzing environment- and sensor-aware robustness, enabling side-by-side evaluation of per-frame adaptive-sensing policies against traditional data-centric training strategies. 

\subsection{Environmental and Sensor Variations}

\noindent\textbf{Ambient-illumination sweep.} 
Real deployments often encounter varying lighting conditions, such as warehouse aisles, outdoor loading bays, and dim indoor environments. To reproduce these diverse conditions in a controlled manner, we built a darkroom enclosed with black curtains to block external light completely. Within this setup, we equipped three Philips Hue White \& Color Ambiance bulbs for precise and programmable brightness control. 
Each bulb was calibrated to a consistent color temperature of 6535 K, with brightness systematically varied across five levels: 5\%, 25\%, 50\%, 75\% and 100\%. This controlled setting ensures uniform illumination conditions, enabling accurate comparisons of RGB gain and exposure settings under identical photon flux.

\noindent\textbf{RGB-sensor parameters.} 
 Most RGB cameras permit manual control of two key RGB sensor parameters: exposure, determining the duration of sensor exposure to incoming light, and gain, which amplifies the captured signal strength.
\begin{itemize}[leftmargin=*,noitemsep,topsep=0pt]
    \item \textbf{Exposure grid.} The camera supports exposure values ranging from 1 $\mu$s to 10,000 $\mu$s. Empirical tests indicated that exposure times below 1,000 $\mu$s yield image quality comparable to the camera's auto-exposure setting in well-lit environments. Given that this threshold is relatively low within the configurable range, we chose 13 logarithmically spaced exposure values to maximize coverage across practical operating conditions. 
    
    \item \textbf{Gain grid.} The RGB gain of the camera ranges from 0 to 128. Through visual inspection, we found that linear steps produce more diverse characteristics than exponential spacing. Therefore, we selected 9 linearly spaced gain settings spanning the full range.
\end{itemize}

\noindent\textbf{Depth capture modes.} 
Depth maps are generated via stereo matching, estimating depth by aligning left and right camera images.
To control the depth sensing behavior, we leverage four predefined depth presets provided by Intel, each optimized for different density–accuracy characteristics~\cite{realsense-preset}:

\begin{itemize}[leftmargin=*,noitemsep,topsep=0pt]
    \item \textbf{Default} offers clean depth maps with reduced noise and well-defined edges, for general-purpose applications.
    
    \item \textbf{High Accuracy} provides highly reliable depth estimates by enforcing stricter confidence thresholds, albeit at the expense of a lower fill rate.

    \item \textbf{High Density} increases the fill factor, providing more complete depth maps, particularly effective for detecting more surfaces in low-texture areas.

    \item \textbf{Medium Density} offers a balanced trade-off between fill factor and accuracy, aiming to balance depth map completeness and precision.
\end{itemize}

\subsection{Dataset Collection}

To capture each scene systematically, we kept the object and camera fixed across all sensor and brightness configurations in a given split. To reduce acquisition time, we distributed the depth capture modes across RGB sensor settings rather than capturing every RGB–depth configuration pair. With the object and camera held fixed, RGB-D inputs can be formed by independently selecting RGB images and depth maps. Scenes were assigned exclusively to specific splits, ensuring that poses and RGB sensor configurations do not overlap between training and test sets. 
During acquisition, we used the Intel RealSense SDK to align each depth frame with its corresponding RGB frame and retrieve intrinsic.
To capture diverse object poses, we used a motorized turntable to vary object orientation and adjusted the camera's position and tilt angle. Each object was rigidly mounted onto a ChArUco board, which was firmly placed on the turntable.

For automatic collection, the turntable was remotely controlled, and deliberate delays were introduced after changing lighting brightness or sensor parameters for sufficient stabilization. 
We employed one smart IR remote-control hub to send commands to the turntable and another hub to control three smart bulbs. Both hubs received commands from a cloud-based server, which coordinated the entire data-capture sequence via an integrated control script (\cref{fig:testbed}).

\subsection{Data Processing}

\noindent\textbf{Data validation.}
We visually inspected all images across each parameter configuration, and excluded any scene appeared inconsistent due to lighting or sensor adjustment delays. Each excluded scene was removed from all brightness–sensor combinations.

\noindent\textbf{Stabilizing auto-exposure frames.} 
The default setting to capture RGB images is the auto-exposure mode.
For the D455 camera, we observed that the auto-exposure setting depends on the gain value set in the previous frame, leading to frame-to-frame brightness jitter. To suppress this effect, we captured successive auto-exposure shots of each scene (5 shots for Train-Def/Var and 4 shots for Test-Def/Var), each preceded by a different gain initialization, and averaged them to obtain a stable reference image.

\noindent\textbf{Ground-Truth annotation.}
For each pose, we used the 50\%-brightness, averaged auto-exposure RGB frame and went through the annotation process below.
\begin{itemize}[leftmargin=*,noitemsep,topsep=0pt]
    \item \textbf{Object masks:} Rough bounding boxes were generated using Grounding DINO~\cite{groundingdino}, guided by shape-aware object-specific textual prompts. Boxes were manually reviewed and corrected if needed. The refined boxes were passed to FastSAM~\cite{fastsam} to generate precise instance-level masks.

    \item \textbf{Object poses:} An initial camera-to-object transformation was provided by detecting the ChArUco markerboard on the turntable. The object poses were then refined precisely in 3D using the BOP annotation tool~\cite{ann_tool} with depth maps captured in High Density mode. Lastly, all refined poses were re-projected onto the RGB images and visually inspected to confirm alignment accuracy.

    \item \textbf{Accuracy of GT annotation:} Following the protocol in~\cite{tless, hb}, we calculate the difference $\delta = d_c - d_r$, where $d_c$ and $d_r$ denote the captured and rendered depth map, respectively. The rendered depth map is generated using the GT pose. Differences over 5 cm are treated as outliers and omitted. Using the high-accuracy depth sensing mode, the resulting depth differences exhibit a mean of $\mu_{|\delta|} = $ 2.32 mm and a standard deviation of $\sigma_\delta = $ 3.29 mm.
    
\end{itemize}

\section{Experiments} 

\noindent\textbf{Goal.}  \ours is designed to isolate an under-measured robustness axis---illumination and sensor-setting variation---and to make its effects measurable across 6D pose estimation methods. Accordingly, our experiments quantify (i) how much accuracy varies under these shifts, and (ii) the headroom available if one could select sensor settings effectively at test time (used here as diagnostic upper bounds).

\noindent\textbf{Models.} 
To quantify the effect of illumination and sensor variations---and the gains from sensor adaptation---relative to conventional training–inference paradigms (large‑scale pretraining and standard supervised pipelines), we evaluate several state‑of‑the‑art methods: 
\begin{itemize}[leftmargin=*]
    \item \textbf{Pretrained, generalizable models (unseen objects):} \textbf{GigaPose}~\cite{gigapose}, \textbf{SAM-6D}~\cite{sam6d}, and \textbf{FoundationPose}~\cite{foundationpose}. 
    GigaPose and SAM-6D were pretrained on the large-scale MegaPose~\cite{megapose}, which contains objects from ShapeNet~\cite{shapenet} and GSO~\cite{gso}. FoundationPose uses a custom data-generation pipeline with 3D assets from GSO and Objaverse~\cite{objaverse}. 

    \item \textbf{Instance-level models (known objects):} \textbf{ZebraPose}~\cite{zebrapose}, an RGB-based method with a ResNet-34 backbone; \textbf{GDRNPP}~\cite{gdrnpp}, which builds on GDR-Net~\cite{gdrnet} using a ConvNeXt~\cite{convnext} backbone and depth-based refinement module; and \textbf{HiPose}~\cite{hipose}, an RGB-D approach that fuses ConvNeXt-based RGB features with RandLA-Net~\cite{randla} depth features.
    These models are trained on \ours. 
\end{itemize}

\begin{table}[t]
  \caption{\textbf{AUC@[0:0.1] performance of various sensor control methods for unseen object pose estimation models} }
  \label{tab:unseen-comparison}
  \centering
    \vspace{-2ex}
      \begin{subtable}[t]{\linewidth}
    \centering
    \caption{GigaPose~\cite{gigapose}}
    
    \label{tab:gigapose}
    \resizebox{\linewidth}{!}{
    \begin{tabular}{l c c c c c}
    \toprule 
    \textbf{Object} & \textbf{Baseline} & \textbf{Depth-Only} & \textbf{RGB-Only} & \textbf{Oracle-F} & \textbf{Oracle-D} \\
    \midrule
    \multirow{2}{*}{Spray} &    
\multirow{2}{*}{64.69} &
    72.01 & 75.21 & 67.83 & 79.31 \\
    &    & (+7.32) & (+10.52) & (+3.14) & (+14.62) \\
    \multirow{2}{*}{Pringles} &    
    \multirow{2}{*}{21.94} &
    44.01 & 45.50 & 29.72 & 68.95 \\
    &    & (+22.07) & (+23.56) & (+7.78) & (+47.01) \\
    \multirow{2}{*}{Tincase} &    
    \multirow{2}{*}{74.03} &
    81.38 & 77.36 & 75.18 & 83.30 \\
    &    & (+7.35) & (+3.33) & (+1.15) & (+9.27) \\
    \multirow{2}{*}{Sandwich} &    
    \multirow{2}{*}{65.35} &
    72.94 & 75.86 & 65.81 & 84.05 \\
    &    & (+7.59) & (+10.51) & (+0.46) & (+18.70) \\
    \multirow{2}{*}{Mouse} &    
    \multirow{2}{*}{54.99} &
    65.80 & 66.52 & 51.71 & 75.32 \\
    &    & (+10.81) & (+11.53) & (-3.28) & (+20.33) \\
    \multirow{2}{*}{Duck} &    
    \multirow{2}{*}{72.15} &
    78.78 & 77.44 & 72.34 & 82.64 \\
    &    & (+6.63) & (+5.29) & (+0.19) & (+10.49) \\
    \midrule
    \multirow{2}{*}{Overall} &    
    \multirow{2}{*}{58.86} &
     69.15 & 69.65 & 60.43 & 78.93 \\
     &    & (+10.29) & (+10.79) & (+1.57) & (+20.07) \\    \bottomrule
    \end{tabular}}
  \end{subtable}

  \begin{subtable}[t]{\linewidth}
    \centering
    \caption{SAM-6D~\cite{sam6d}}
    \label{tab:sam6d}
    \resizebox{\linewidth}{!}{
        \begin{tabular}{l c c c c c}
    \toprule 
    \textbf{Object} & \textbf{Baseline} & \textbf{Depth-Only} & \textbf{RGB-Only} & \textbf{Oracle-F} & \textbf{Oracle-D} \\
    \midrule
    \multirow{2}{*}{Spray} &    
    \multirow{2}{*}{83.14} &
    85.62 & 82.64 & 83.34 & 88.50 \\
    &    & (+2.48) & (-0.50) & (+0.20) & (+5.36) \\
    \multirow{2}{*}{Pringles} &    
    \multirow{2}{*}{64.09} &
    73.96 & 74.07 & 64.54 & 80.63 \\
    &    & (+9.87) & (+9.98) & (+0.45) & (+16.54) \\
    \multirow{2}{*}{Tincase} &    
    \multirow{2}{*}{73.42} &
    77.53 & 80.19 & 68.95 & 83.32 \\
    &    & (+4.11) & (+6.77) & (-4.47) & (+9.90) \\
    \multirow{2}{*}{Sandwich} &    
    \multirow{2}{*}{78.39} &
    78.99 & 80.02 & 78.38 & 80.81 \\
    &    & (+0.60) & (+1.63) & (-0.02) & (+2.41) \\
    \multirow{2}{*}{Mouse} &    
    \multirow{2}{*}{55.72} &
    62.25 & 67.69 & 58.63 & 72.45 \\
    &    & (+6.53) & (+11.96) & (+2.91) & (+16.72) \\
    \multirow{2}{*}{Duck} &    
    \multirow{2}{*}{76.67} &
    82.42 & 83.40 & 78.58 & 87.64 \\
    &    & (+5.74) & (+6.72) & (+1.90) & (+10.97) \\
    \midrule
    \multirow{2}{*}{Overall} &    
    \multirow{2}{*}{71.91} &
     76.80 & 78.00 & 72.07 & 82.22 \\
     &    & (+4.89) & (+6.09) & (+0.16) & (+10.32) \\
    \bottomrule
    \end{tabular}}
  \end{subtable}
  \begin{subtable}[t]{\linewidth}
    \centering
    \caption{FoundationPose~\cite{foundationpose}}
    \label{tab:foundationpose}
    \resizebox{\linewidth}{!}{
    \begin{tabular}{l c c c c c}
    \toprule 
    \textbf{Object} & \textbf{Baseline} & \textbf{Depth-Only} & \textbf{RGB-Only} & \textbf{Oracle-F} & \textbf{Oracle-D} \\
    \midrule
    \multirow{2}{*}{Spray} &    
    \multirow{2}{*}{84.63} &
    86.52 & 87.60 & 83.96 & 89.01 \\
    &    & (+ 1.89) & (+2.97) & (-0.67) & (+4.38)\\
    \multirow{2}{*}{Pringles} &    
    \multirow{2}{*}{34.26} &
    50.72 & 50.78 & 37.73 & 74.69 \\
    &    & (+16.46) & (+16.52) & (+3.47) & (+40.43) \\
    \multirow{2}{*}{Tincase} &    
    \multirow{2}{*}{38.24} &
    56.16 & 69.38 & 40.07 & 82.79 \\
    &    & (+17.92) & (+31.14) & (+1.83) & (+44.55) \\
    \multirow{2}{*}{Sandwich} &    
    \multirow{2}{*}{83.17} &
    86.40 & 85.12 & 83.14 & 88.11 \\
    &    & (+3.23) & (+1.95) & (-0.03) & (+4.94) \\
    \multirow{2}{*}{Mouse} &    
    \multirow{2}{*}{70.22} &
    74.34 & 75.86 & 70.69 & 79.98 \\
    &    & (+4.12) & (+5.64) & (+0.46) & (+9.75)  \\
    \multirow{2}{*}{Duck} &    
    \multirow{2}{*}{84.78} &
    86.87 & 89.15 & 84.72 & 90.67 \\
    &    & (+2.09) & (+4.37) & (-0.06) & (+5.89) \\
    \midrule
    \multirow{2}{*}{Overall} &    
    \multirow{2}{*}{65.88} &
     73.50 & 76.32 & 66.72 & 84.21 \\
     &    & (+7.62) & (+10.43) & (+0.83) & (+18.32) \\
    \bottomrule
    \end{tabular}}
  \end{subtable}
  \vspace{-2.5ex}
\end{table}

\noindent\textbf{Sensor-setting evaluation protocols.} 
To characterize sensitivity and headroom, we report five regimes:  
(1) \textbf{Baseline:} default settings for both RGB (e.g., auto-exposure) and depth sensors; 
(2) \textbf{Depth-Only (Oracle):} per-scene best depth preset among the available modes and auto-exposure for RGB;
(3) \textbf{RGB-Only (Oracle):} per-scene best exposure/gain, keeping default depth capture mode; 
(4) \textbf{Oracle-Fixed (Oracle-F):} a single globally best RGB-D configuration applied to all test scenes;
(5) \textbf{Oracle-Dynamic (Oracle-D):} per-scene best RGB-D configuration.
Here, ``Oracle'' represents an \textit{offline upper bound} obtained by evaluating the full sensor-configuration grid and marking a scene correct if \textit{any} configuration yields a correct pose.

\noindent\textbf{Metrics.} 
We use the standard Average Distance (ADD) metric~\cite{LM}, which measures the average point-wise distance between model points transformed by the predicted pose and the ground-truth pose. Based on ADD, we report two evaluation measures: \textbf{AUC@[0:0.1]}, which corresponds to the area under the ADD recall curve up to a threshold of 10\% of the object diameter, and \textbf{AR@5}, which denotes the recall at a threshold of 5\% of the object diameter.
Additional metrics---average recall of VSD, MSSD and MSPD~\cite{tudl}---are reported in Appendix~\cref{appendix:bop-metrics}.

\subsection{Analysis on Pretrained, Generalizable  Pose Estimation Models}
\label{sec:unseen}

Generalizable 6D pose estimators leverage large-scale pretraining to handle clutter, occlusion, and novel objects. \ours removes these factors---single object, clean background, no occlusion---so that evaluation isolates robustness to \textbf{environmental and sensor shifts}, testing whether pretrained robustness transfers to this \textit{orthogonal and underexplored axis}: variations in illumination and sensor parameters that are rarely addressed in existing benchmarks.

\cref{tab:unseen-comparison} presents the impact of illumination and sensor variations on 6D pose estimation, given GT object segmentations as input. We provide the end-to-end evaluation results on CNOS~\cite{cnos} segmentation and the full SAM-6D pipeline in Appendix ~\cref{appendix:cnos-pretrained,appendix:sam-6d-full}.

\begin{figure*}[t]
     \centering
     \includegraphics[width=\textwidth]{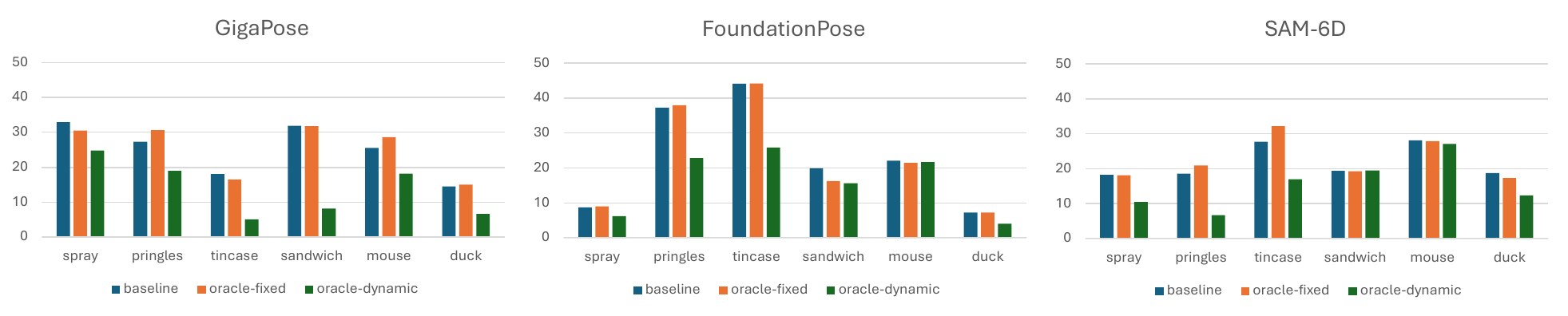}
     \vspace{-5.5ex}
     \caption{\textbf{Overall standard deviation of AUC computed across all brightness and scenes evaluated under Baseline, Oracle-Fixed, Oracle-Dynamic.}}
     \vspace{-2ex}
     \label{fig:scene_std}
\end{figure*}

\noindent\textbf{Sensitivity under default sensing.} 
Under Baseline, all pretrained models exhibit significant performance drop. 
This indicates that strong pretraining alone does not automatically extend to sensor- and illumination-induced shifts.

\noindent\textbf{Headroom diagnostics via oracle selection.} 
Oracle regimes reveal substantial headroom of test-time sensor control without model retraining. Averaged across the three pretrained models, RGB‑Only improves AUC by +9.1 pp and Depth-Only by +7.6 pp over Baseline, indicating that defaults are not model-optimal. 
Furthermore, Oracle-D achieves the best results, with +16.2 pp over Baseline and +7.1 pp over the second-best RGB-Only. This gap indicates that RGB and depth provide \textit{complementary cues} and that their optimal settings can interact non-trivially. 

\noindent\textbf{Limitation of fixed tuning.}  
In contrast, Oracle-F provides only +0.9 pp on average---and can even underperform Baseline (e.g., SAM-6D on Tincase). This finding shows that no single sensor configuration generalizes across lighting conditions and object appearances, motivating evaluation and methods to be \textbf{scene-adaptive} rather than globally tuned.

\noindent\textbf{Stability and object-wise disparity.} 
\ours reveals substantial object-wise performance gaps under Baseline: the average gap between the best‑ and worst‑performing objects is 43.3 pp, shrinking to 15.7 pp under Oracle-D. Per-scene oracle selection also reduces \textit{scene‑level variability} within each object, lowering the overall AUC standard deviation by 8.3 pp relative to Baseline (\cref{fig:scene_std}).

Together, these results show that sensor/illumination shifts are a distinct robustness axis that remains challenging even for pretrained generalizable pose estimators. Qualitative examples are provided in Appendix~\cref{appendix: qualitative-unseen}.

\subsection{Analysis on Instance-Level Pose Estimation Models}
\label{sec:instance-level}
\vspace{-1ex}
Instance‑level methods follow the standard train–test setup in which the training and test sets share the same object domain. Building on \cref{sec:unseen}, our goal here is twofold: (i) to test whether environmental and sensor variations remain challenging without the unseen‑object factor, and (ii) to assess whether adapting sensor configuration at test time remains effective even when the training data already includes augmentations or physical sensor and illumination variation.

For training, we generated 10k photorealistic (PBR) synthetic images per object using BlenderProc~\cite{blenderproc}. Each model was trained independently on a per-object basis under three schemes: (1) \textbf{PBR + Train-Def} (default physical images only), (2)  \textbf{PBR + Train-Def with augmentation} (including linear contrast adjustment, brightness enhancement and others; see Appendix~\cref{appendix:augmentation}.
), and (3) \textbf{PBR + Train-Var} (full sensor and lighting variation). In all experiments, we excluded refinement modules for ZebraPose and GDRNPP, and used the full pipeline for HiPose. 
Since HiPose is the only RGB-D method, we report results in two parts.

\cref{tab:pose-comparison} evaluates the impact of illumination and RGB sensor variations (with the default depth preset) given GT object mask. We consider three sensor-control settings
: \textbf{Auto} (auto-exposure), \textbf{Rand} (average performance across randomly selected exposure-gain pairs), and \textbf{Oracle} (per-scene upper bound over the full RGB grid). 
\cref{tab:multimodal-comparion} evaluates HiPose (training scheme 1) on Test-Var under the multimodal regimes introduced in \cref{sec:unseen}.

\begin{table*}[t]
  \caption{\textbf{AR@5 performance of various sensor control methods under different training and test settings of \ours for instance-level pose estimation models: ZebraPose, GDRNPP, HiPose.}}
  \label{tab:pose-comparison}
  \vspace{-2ex}
  \centering
  \resizebox{\linewidth}{!}{
  \begin{tabular}{ll cccc cccc cccc}
    \toprule
    \multicolumn{2}{c}{} 
    & \multicolumn{4}{c}{\textbf{ZebraPose~\cite{zebrapose}}}
    & \multicolumn{4}{c}{\textbf{GDRNPP~\cite{gdrnpp}}}
    & \multicolumn{4}{c}{\textbf{HiPose~\cite{hipose}}} \\
    \cmidrule(lr){3-6} \cmidrule(lr){7-10} \cmidrule(lr){11-14}
    \multicolumn{2}{c}{} 
    & \textbf{Test-Def} & \multicolumn{3}{c}{\textbf{Test-Var}}
    & \textbf{Test-Def} & \multicolumn{3}{c}{\textbf{Test-Var}}
    & \textbf{Test-Def} & \multicolumn{3}{c}{\textbf{Test-Var}} \\
    \cmidrule(lr){3-6} \cmidrule(lr){7-10} \cmidrule(lr){11-14}
    \textbf{Object} & \textbf{Train} 
      & Auto & Auto & Rand & Oracle 
      & Auto & Auto & Rand & Oracle 
      & Auto & Auto & Rand & Oracle \\
    \midrule

    \multirow{3}{*}{Spray}
      & PBR + Train-Def
        & 100.0 & 96.10 & 54.61 & 98.54
        & 100.0 & 96.59 & 52.02 & 99.02
        & 100.0 & 99.51 & 59.88 & 100.0 \\
      & PBR + Train-Def w/ Aug
        & 95.12 & 93.66 & 64.98 & 99.51
        & 100.0 & 98.54 & 71.88 & 100.0
        & 100.0 & 100.0 & 89.76 & 100.0 \\
      & PBR + Train-Var
        & 97.56 & 97.07 & 79.24 & 97.56
        & 97.56 & 97.56 & 85.93 & 98.54
        & 100.0 & 100.0 & 100.0 & 100.0 \\

    \midrule
    \multirow{3}{*}{Pringles}
      & PBR + Train-Def
        & 90.48 & 84.29 & 43.31 & 95.71
        & 69.05 & 60.95 & 35.64 & 89.05
        & 100.0 & 99.52 & 58.71 & 100.0 \\
      & PBR + Train-Def w/ Aug
        & 64.29 & 64.76 & 44.38 & 93.33
        & 80.95 & 79.05 & 49.17 & 89.52
        & 97.62 & 97.62 & 72.95 & 100.0 \\
      & PBR + Train-Var
        & 83.33 & 82.38 & 64.86 & 90.95
        & 90.48 & 90.00 & 72.21 & 97.14
        & 97.62 & 97.62 & 86.88 & 98.10 \\

    \midrule
    \multirow{3}{*}{Tincase}
      & PBR + Train-Def
        & 97.78 & 86.67 & 45.78 & 96.44
        & 91.11 & 81.33 & 46.56 & 92.89
        & 97.78 & 86.67 & 53.62 & 96.44  \\
      & PBR + Train-Def w/ Aug
        & 97.78 & 94.67 & 59.38 & 100.0        
        & 82.22 & 82.22 & 55.78 & 95.11
        & 97.78 & 96.89 & 86.62 & 97.78 \\
      & PBR + Train-Var
        & 97.78 & 98.22 & 75.09 & 100.0
        & 100.0 & 99.56 & 82.00 &  100.0
        & 97.78 & 97.78 & 97.76 & 97.78 \\

    \midrule
    \multirow{3}{*}{Sandwich}
      & PBR + Train-Def
        & 94.87 & 89.23 & 48.38 & 99.49
        & 97.44 & 88.21 & 47.03 & 96.41 
        & 94.87 & 92.87 & 57.21 & 94.87 \\
      & PBR + Train-Def w/ Aug
        & 92.31 & 88.72 & 57.82 & 99.49
        & 94.87 & 92.31 & 63.72 & 98.97  
        & 94.87 & 94.36 & 84.46 & 94.87 \\
      & PBR + Train-Var
        & 92.31 & 90.77 & 67.03 & 96.41
        & 97.44 & 97.44 & 82.64 & 100.0
        & 94.87 & 94.87 & 94.87 & 94.87 \\

    \midrule
    \multirow{3}{*}{Mouse}
      & PBR + Train-Def
        & 70.73 & 62.44 & 27.00 & 87.80
        & 95.12 & 76.10 & 31.44 & 88.78 
        & 90.24 & 79.02 & 43.68 & 87.80 \\
      & PBR + Train-Def w/ Aug
        & 48.78 & 48.78 & 32.46 & 87.80
        & 78.05 & 72.68 & 43.29 & 86.83 
        & 90.24 & 90.24 & 82.44 & 90.73 \\
      & PBR + Train-Var
        & 46.34 & 47.80 & 35.07 & 71.22
        & 80.49 & 80.49 & 66.39 & 91.22 
        & 90.24 & 90.24 & 87.29 & 90.73 \\

    \midrule
    \multirow{3}{*}{Duck}
      & PBR + Train-Def
        & 93.02 & 84.19 & 28.95 & 94.88
        & 95.35 & 84.65 & 39.63 & 99.07  
        & 97.67 & 80.00 & 44.56 & 99.53
 \\
      & PBR + Train-Def w/ Aug
        & 93.02 & 84.65 & 54.63 & 98.60
        & 90.70 & 84.65 & 57.91 & 99.07
        & 97.67 & 98.60 & 82.86 & 100.0 \\
      & PBR + Train-Var
        & 93.02 & 93.02 & 71.77 & 98.60
        & 97.67 & 97.67 & 81.44 & 99.53
        & 97.67 & 98.60 & 97.49 & 100.0 \\

    \midrule
    \multirow{3}{*}{Overall}
      & PBR + Train-Def
        & 91.15 & 83.82 & 41.34 & 95.48
        & 91.34 & 81.30 & 42.05 & 94.20
        & 96.76 & 89.59 & 52.94 & 96.44 \\
      & PBR + Train-Def w/ Aug
        & 81.88 & 79.21 & 52.27 & 96.46
        & 87.80 & 84.91 & 56.96 & 94.92
        & 96.36 & 96.29 & 83.18 & 97.23 \\
      & PBR + Train-Var
        & 85.06 & 84.88 & 65.51 & 92.46
        & 93.94 & 93.79 & 78.44 & 97.74
        & 96.36 & 96.52 & 94.05 & 96.91 \\

    \bottomrule
  \end{tabular}}
  \vspace{-3ex}
\end{table*}

\noindent\textbf{RGB/illumination shift remains a challenge, even for known objects.} 
\cref{tab:pose-comparison} shows that when trained with Train-Def, Auto performance consistently drops from Test-Def to Test-Var (e.g., overall -7.3 pp for ZebraPose, -10.0 pp for GDRNPP, -7.2 pp for HiPose), confirming that sensor/illumination shifts are impactful even with identical objects and backgrounds. 
Oracle selection (RGB) substantially closes this gap: Oracle on Test-Var consistently outperforms Auto on Test-Var and often matches or exceeds Auto on Test-Def, indicating that a significant portion of the failure is attributable to \textbf{capture conditions that suppress or corrupt informative cues}.

\noindent\textbf{Multimodal shifts for an instance‑level RGB‑D model.} 
HiPose performance in \cref{tab:multimodal-comparion} reveals that, while all test-time sensor‑selection regimes improve over Baseline, the average gain from RGB‑Only (+8.02 pp) is substantially larger than that from Depth‑Only (+2.44 pp), unlike the pretrained setting where both modalities often contribute strongly (\cref{tab:unseen-comparison}). This is expected because HiPose is trained instance‑level with depth maps from the same domain as the Baseline depth preset, and depth is generally less sensitive to illumination changes than RGB. 
Importantly, depth‑mode selection still provides performance gains, and joint per‑scene selection remains best overall (Oracle‑Dynamic: +10.07 pp), indicating that RGB and depth can still provide complementary robustness depending on the scene.

\begin{table}[t]
\centering
\caption{\textbf{Overall AUC@[0:0.1] performance of various sensor control methods for HiPose, trained on PBR + Train-Def and tested on Test-Var.}}
\vspace{-2ex}
\label{tab:multimodal-comparion}
\resizebox{\linewidth}{!}{
\begin{tabular}{l c c c c c}
\toprule 
 & 
\makecell[c]{\textbf{Baseline}} &
\makecell[c]{\textbf{Depth-Only}} &
\makecell[c]{\textbf{RGB-Only}} &
\makecell[c]{\textbf{Oracle-F}} &
\makecell[c]{\textbf{Oracle-D}} \\
\midrule
\multirow{2}{*}{Overall} &
\multirow{2}{*}{74.66} &
77.22 & 82.68 & 64.74 & 84.73 \\
& 
& (+2.44) & (+8.02) & (-9.92) & (+10.07)  \\
\bottomrule
\end{tabular}}
\vspace{-3ex}
\end{table}

\noindent\textbf{Data-centric robustness is helpful but incomplete.} 
In \cref{tab:pose-comparison}, data augmentation generally improves Auto on Test‑Var for GDRNPP/HiPose, but can hurt for some cases (e.g., Mouse for GDRNPP), indicating that synthetic perturbations are not uniformly reliable. 
Meanwhile, training on expanded physically captured variation (PBR + Train‑Var, 3.5$\times$ larger than PBR + Train‑Def) yields larger and more consistent gains than augmentation overall. 
However, it can still fail in difficult regimes (e.g., Mouse for ZebraPose), suggesting that ``larger and more diverse training data'' is not a guaranteed solution when the sensing process itself removes critical cues.

\noindent\textbf{Sensor-aware diagnostics vs. scaling data.} 
An important finding is that oracle selection with limited training can rival or exceed data-scaling baselines: for example, ZebraPose trained on Train‑Def with Oracle achieves $>$+10 pp higher accuracy on Test‑Var than Auto trained on Train‑Var (\cref{tab:pose-comparison}). 
This highlights the benchmark's core message: robustness under sensor/environment shifts is not only a modeling problem, but also a sensing problem---capturing more reliable observations can be as important as scaling training data.
Qualitative examples are provided in Appendix~\cref{appendix:qualitative-instance}.


\begin{table}[t]
  \caption{\textbf{Performance comparison under different sensor selection strategies.}
Values in parentheses denote ADD-S variant. Supervised models are trained under the PBR + Train-Def setting.}
  \label{tab:proxy_performance}
  \centering
    \vspace{-2ex}
  \begin{subtable}[t]{\linewidth}
    \centering
    \caption{Supervised Models (AR@5)}
    \label{tab:proxy_supervised}
    \resizebox{0.8\linewidth}{!}{
    \begin{tabular}{l cccc}
      \toprule
      \multicolumn{1}{c}{}
      & \textbf{Test-Def} & \multicolumn{3}{c}{\textbf{Test-Var}} \\
      \cmidrule(lr){2-2}\cmidrule(lr){3-5}
      \textbf{Model}
        & Auto & Auto & Oracle & Proxy \\
      \midrule
      \multirow{2}{*}{ZebraPose~\cite{zebrapose}}
         & 88.75 &  80.73  & 94.75 & 91.16 \\
         & (98.38) &  (93.97)  & (98.48) & (97.99) \\
      \multirow{2}{*}{GDRNPP~\cite{gdrnpp}}
         & 91.34 & 80.99 & 93.81 &  90.12 \\
         & (99.21) &  (94.22)  & (98.57) & (98.40) \\
      \multirow{2}{*}{HiPose~\cite{hipose}}
         & 94.75 &  86.85  & 96.46 &  95.51 \\
         & (95.52) & (88.82)  &  (97.27) &  (97.19) \\
      \bottomrule
    \end{tabular}}
  \end{subtable}
  \begin{subtable}[t]{\linewidth}
    \centering
    \caption{Pretrained Models (AUC@[0:0.1])}
    \label{tab:proxy_pretrained}
    \resizebox{0.8\linewidth}{!}{
    \begin{tabular}{l cccc}
      \toprule
      \textbf{Model}
        & Baseline & Oracle & Proxy \\
      \midrule
      \multirow{2}{*}{GigaPose~\cite{gigapose}}      
          & 58.55 & 78.28 & 64.84 \\
          & (78.24) & (88.92) & (86.18) \\
      \multirow{2}{*}{SAM-6D~\cite{sam6d}}
          & 63.75 & 81.36 & 71.03 \\
          & (77.89) & (88.10) &  (85.88) \\
      \multirow{2}{*}{FoundationPose~\cite{foundationpose}}
          & 62.95 & 83.50 & 64.25 \\
          & (83.48) & (90.54) & (87.89) \\
      \bottomrule
    \end{tabular}}
  \end{subtable}
  \vspace{-3ex}
\end{table}

\subsection{Towards Deployable Sensor Adaptation} 

Oracle selection quantifies headroom but is not deployable, since GT pose is unavailable at test time. To ground future work, we evaluate a simple proxy-based baseline that attempts to pick sensor settings as a practical baseline.

\noindent\textbf{Proxy design.} 
We consider VSD (Visible Surface Discrepancy)~\cite{tudl} as a proxy for test-time sensor selection.
Since VSD captures the geometric consistency between the ground-truth and estimated poses, it provides a potential signal for assessing pose quality and guiding sensor adaptation.
For each scene, we analyze the correlation between ADD and VSD across different sensor configurations.
The results (Appendix~\cref{appendix:add-vsd-corr}) show moderate-to-strong correlation (above 0.86 for supervised models, 0.62 for pretrained models).

With the motivation, we design a VSD-like proxy metric: 
{\small{
\begin{equation}
\label{eq:proxy_vsd}
\begin{aligned}
&\operatorname{proxy}_{\mathrm{VSD}}(\hat{S}, S_I, \hat{V}, V_I, \tau)
\\
&\quad =
\underset{p \in \hat{V} \cup V_I}{\operatorname{avg}}
\begin{cases} 0, & \text{if } p \in \hat{V} \cap V_I
\land |\hat{S}(p)-S_I(p)| < \tau, 
\\
1, & \text{otherwise.}
\end{cases}
\end{aligned}
\end{equation}
}}

\noindent Here $\hat{S}$ and $S_I$ denote the distance maps rendered from the estimated pose and computed from the observed depth map, respectively.
$V_I$ is the object mask, while $\hat{V}$ is the estimated visibility mask obtained from $(\hat{S}, S_I)$.
We compute the IoU between $V_I$ and $\hat{V}$, and use it as a tie-breaking criterion to resolve ambiguous cases during sensor selection.

\noindent\textbf{Results.} 
\cref{tab:proxy_performance} evaluates the proxy-based sensor selection using CNOS-generated masks. For supervised/instance-level models on Test-Var (\cref{tab:proxy_supervised}), the proxy consistently improves over Auto and can approach the oracle ceiling in favorable cases (e.g., HiPose).
For pretrained generalizable models (\cref{tab:proxy_pretrained}), the proxy improves over Baseline with only marginal gain for FoundationPose. The larger gain for FoundationPose under ADD-S suggests that \textbf{symmetry and pose ambiguity} and the proxy's lack of RGB appearance cues may contribute to the limited gain. 
Overall, this experiment reinforces the benchmark's motivation: \ours exposes large headroom, but achieving it in practice requires stronger, multimodal proxy signals and principled sensor selection strategies.
\section{Conclusion}

We introduced \ours, a physically captured RGB-D benchmark that \emph{systematically sweeps} illumination, RGB exposure/gain, and depth presets to isolate the impact of environment and sensor shifts on object pose estimation.
\ours provides accurate pose ground truth, standardized splits, and a dense grid of real sensor-induced artifacts that are difficult to reproduce with synthetic augmentation.

Using this benchmark, we establish \textbf{sensor/environment shift as a distinct
robustness axis} that is rarely addressed by existing pose benchmarks or by large-scale pretraining alone. 
Across three pretrained, generalizable estimators and three instance-level estimators, performance varies substantially with illumination and sensor settings even in uncluttered single-object scenes.
Moreover, under pretrained models, per-scene sensing adaptation reveals substantial headroom, improving accuracy by +16.2 pp over defaults, whereas a single globally fixed configuration yields only marginal gains. 
Beyond accuracy, per-scene adaptation also reduces object-wise disparity and scene-level variability, indicating improved \emph{stability} in addition to higher accuracy.
Finally, to illustrate the opportunities enabled by \ours, we benchmark simple sensor-selection baselines alongside oracle upper bounds. 
While the oracle reveals large headroom, a consistency-based strategy closes only a small fraction of the gap for pretrained models, highlighting the need for future research on reliable, model-aware selection signals that jointly leverage photometric and geometric cues.

\noindent\textbf{Limitations and Future Work.} 
\ours currently focuses on single-object scenes involving only six objects, and varies illumination primarily through brightness magnitude.
Moreover, visible ChArUco boards may introduce artifacts not encountered during real deployment (Appendix~\cref{appendix:markerless}). Thus, the generalizability of our findings to more complex real-world environments remains to be investigated.
Future work will expand (i) \textbf{scene complexity} (multi-object occlusions and diverse backgrounds), (ii) \textbf{illumination variation}
(directionality and mixed color temperatures), and (iii)
\textbf{annotation strategies} that avoid visible markers.
Beyond dataset expansion, \ours motivates new algorithmic directions: reliable practical proxy, 
learning-based or bandit-style sensor-selection policies, multimodal selection
criteria that combine RGB appearance with depth geometry, and real-time sensing
frameworks that adapt capture settings online to maintain robust observations for pose estimation.

\section*{Acknowledgments}

This work was supported in part by the Institute of Information \& communications Technology Planning \& Evaluation (IITP) grant funded by the Korea government (MSIT) (No. RS-2026-25507282, RS-2023-00216821), in part by the National Research Foundation (NRF) of Korea grant funded by the Korea government (MSIT) (No. RS-2023-00222663), and in part by Mid-Career Bridging Program through Seoul National University.

{
    \small
    \bibliographystyle{ieeenat_fullname}
    \bibliography{main}
}

\clearpage
\section*{Appendix}
\addcontentsline{toc}{section}{Appendix}
\appendix

\section{Experimental Details} 
\label{appendix:details}
\subsection{Computing Resources}
All models were trained on a high-performance computing cluster equipped with NVIDIA GPUs and Intel Xeon CPUs. Specifically, \textit{ZebraPose} and \textit{GDRNPP} were trained on nodes with RTX 3090 GPUs (24GB VRAM) and Intel Xeon Silver 4210R CPUs, while \textit{HiPose} was trained on a node with A100 GPUs (80GB VRAM) and an Intel Xeon Gold 6530 CPU. Each training node had access to 512GB of RAM.

\subsection{Pretrained Weights for Unseen Object Pose Estimation Models}
For GigaPose, we used the pretrained weight files `gigaPose\_v1' (coarse) and `refiner-rgbd-288182519' (refiner), following instructions from \href{https://github.com/nv-nguyen/gigapose} {GigaPose GitHub repository}.

For FoundationPose, we used the pretrained weight files `2023-10-28-18-33-37' (refiner) and `2024-01-11-20-02-45' (scorer), following instructions from \href{https://github.com/NVlabs/FoundationPose}{FoundationPose GitHub repository}.

For SAM-6D, we used the pretrained weight file `vit\_h' for SAM and the `sam-6d-pem-base.pth' file for the Pose Estimation Module (PEM), following instructions from \href{https://github.com/JiehongLin/SAM-6D} {SAM-6D Github repository}.

\subsection{Model Training for Instance-Level Pose Estimation Models}
All models were trained with a batch size of 32. \textit{ZebraPose} and \textit{HiPose} were both trained for a total of 38{,}000 iterations using the Adam optimizer, with learning rates of 0.0002 and 0.0001, respectively. \textit{GDRNPP} was trained for 120 epochs using the Ranger optimizer with a learning rate of 8e-4 and weight decay of 0.01. A flat-and-anneal learning rate schedule with cosine annealing was applied, starting at 72\% of the total training epochs. A linear warmup was used for the first 1{,}000 iterations with a warmup factor of 0.001. All learning rate schedules and optimizer settings follow the original implementations of each method.

\section{End-to-end evaluation using CNOS segmentation mask}
\label{appendix:cnos-pretrained}
\cref{tab:cno-pretrained} reports the end-to-end performance of GigaPose, SAM-6D, and FoundationPose using the top-1 segmentation masks generated by CNOS.

\begin{table}[t]
  \caption{\textbf{AUC@[0:0.1] performance of various sensor control methods for unseen object pose estimation models given CNOS mask} }
  \label{tab:cno-pretrained}
  \centering
    \vspace{-2ex}
      \begin{subtable}[t]{\linewidth}
    \centering
    \caption{GigaPose}
    \label{tab:cnos-gigapose}
    \resizebox{\linewidth}{!}{
    \begin{tabular}{l c c c c c}
    \toprule 
    \textbf{Object} & \textbf{Baseline} & \textbf{Depth-Only} & \textbf{RGB-Only} & \textbf{Oracle-F} & \textbf{Oracle-D} \\
    \midrule
    \multirow{2}{*}{Spray} &
    \multirow{2}{*}{64.90} &
    71.97 & 74.29 & 67.82 & 79.00 \\
    & & (+7.07) & (+9.39) & (+2.92) & (+14.09) \\
    \multirow{2}{*}{Pringles} &
    \multirow{2}{*}{21.96} &
    43.12 & 40.06 & 26.95 & 66.00 \\
    & & (+21.16) & (+18.10) & (+5.00) & (+44.04) \\
    \multirow{2}{*}{Tincase} &
    \multirow{2}{*}{72.21} &
    79.29 & 77.26 & 70.03 & 83.20 \\
    & & (+7.08) & (+5.05) & (-2.18) & (+10.99) \\
    \multirow{2}{*}{Sandwich} &
    \multirow{2}{*}{65.36} &
    72.89 & 75.43 & 65.91 & 84.08 \\
    & & (+7.52) & (+10.06) & (+0.55) & (+18.72) \\
    \multirow{2}{*}{Mouse} &
    \multirow{2}{*}{54.51} &
    66.27 & 66.07 & 51.72 & 74.74 \\
    & & (+11.76) & (+11.57) & (-2.79) & (+20.23) \\
    \multirow{2}{*}{Duck} &
    \multirow{2}{*}{72.33} &
    78.93 & 77.22 & 71.80 & 82.69 \\
    & & (+6.60) & (+4.88) & (-0.53) & (+10.35) \\
    \midrule
    \multirow{2}{*}{Overall} &
    \multirow{2}{*}{58.55} &
    68.75 & 68.39 & 59.04 & 78.28 \\
    & & (+10.20) & (+9.84) & (+0.49) & (+19.28) \\
    \bottomrule
    \end{tabular}}
  \end{subtable}

  \begin{subtable}[t]{\linewidth}
    \centering
    \caption{SAM-6D}
    \label{tab:cnos-sam6d}
    \resizebox{\linewidth}{!}{
        \begin{tabular}{l c c c c c}
    \toprule 
    \textbf{Object} & \textbf{Baseline} & \textbf{Depth-Only} & \textbf{RGB-Only} & \textbf{Oracle-F} & \textbf{Oracle-D} \\
    \midrule
    \multirow{2}{*}{Spray} &
    \multirow{2}{*}{74.95} &
    82.01 & 85.51 & 74.90 & 89.02 \\
    & & (+7.06) & (+10.56) & (-0.05) & (+14.07) \\
    \multirow{2}{*}{Pringles} &
    \multirow{2}{*}{54.02} &
    67.05 & 70.78 & 52.20 & 79.36 \\
    & & (+13.03) & (+16.76) & (-1.82) & (+25.34) \\
    \multirow{2}{*}{Tincase} &
    \multirow{2}{*}{64.52} &
    74.36 & 77.93 & 60.80 & 82.72 \\
    & & (+9.84) & (+13.41) & (-3.72) & (+18.20) \\
    \multirow{2}{*}{Sandwich} &
    \multirow{2}{*}{76.72} &
    78.47 & 79.91 & 73.33 & 81.03 \\
    & & (+1.75) & (+3.18) & (-3.9) & (+4.30) \\
    \multirow{2}{*}{Mouse} &
    \multirow{2}{*}{38.62} &
    44.90 & 64.43 & 42.10 & 69.37 \\
    & & (+6.28) & (+25.81) & (+3.48) & (+30.75) \\
    \multirow{2}{*}{Duck} &
    \multirow{2}{*}{73.64} &
    81.16 & 82.25 & 74.34 & 86.62 \\
    & & (+7.52) & (+8.61) & (+0.70) & (+12.98) \\
    \midrule
    \multirow{2}{*}{Overall} &
    \multirow{2}{*}{63.75} &
    71.32 & 76.80 & 62.95 & 81.36 \\
    & & (+7.58) & (+13.05) & (-0.80) & (+17.61) \\
    \bottomrule
    \end{tabular}}
  \end{subtable}
  \begin{subtable}[t]{\linewidth}
    \centering
    \caption{FoundationPose}
    \label{tab:cnos-foundationpose}
    \resizebox{\linewidth}{!}{
    \begin{tabular}{l c c c c c}
    \toprule 
    \textbf{Object} & \textbf{Baseline} & \textbf{Depth-Only} & \textbf{RGB-Only} & \textbf{Oracle-F} & \textbf{Oracle-D} \\
    \midrule
    \multirow{2}{*}{Spray} &
    \multirow{2}{*}{84.11} &
    86.34 & 87.49 & 81.68 & 88.93 \\
    & & (+2.23) & (+3.39) & (-2.43) & (+4.82) \\
    \multirow{2}{*}{Pringles} &
    \multirow{2}{*}{33.73} &
    48.90 & 51.02 & 35.16 & 74.61 \\
    & & (+15.17) & (+17.29) & (+1.40) & (+40.88) \\
    \multirow{2}{*}{Tincase} &
    \multirow{2}{*}{37.15} &
    56.06 & 67.98 & 39.91 & 79.44 \\
    & & (+18.91) & (+30.83) & (+1.76) & (+42.29) \\
    \multirow{2}{*}{Sandwich} &
    \multirow{2}{*}{82.30} &
    85.48 & 85.01 & 79.66 & 88.03 \\
    & & (+3.17) & (+2.71) & (-2.65) & (+5.72) \\
    \multirow{2}{*}{Mouse} &
    \multirow{2}{*}{57.31} &
    61.85 & 75.69 & 59.26 & 80.03 \\
    & & (+4.54) & (+18.38) & (+1.95) & (+22.72) \\
    \multirow{2}{*}{Duck} &
    \multirow{2}{*}{83.10} &
    85.75 & 87.94 & 82.80 & 89.97 \\
    & & (+2.65) & (+4.84) & (-0.31) & (+6.87) \\
    \midrule
    \multirow{2}{*}{Overall} &
    \multirow{2}{*}{62.95} &
    70.73 & 75.86 & 62.90 & 83.50 \\
    & & (+7.78) & (+12.90) & (-0.05) & (+20.50) \\
    \bottomrule
    \end{tabular}}
  \end{subtable}
  \vspace{-2.5ex}
\end{table}


\section{Evaluation of the full SAM-6D pipeline}
\label{appendix:sam-6d-full}
\cref{tab:sam6d-fullpipeline-comparion} presents SAM-6D performance based on the Top-30 segmentation results produced by its ISM.
\begin{table}[H]
\centering{
\caption{\textbf{AUC@[0:0.1] performance of SAM-6D with its Image Segmentation Model (ISM).}}
\label{tab:sam6d-fullpipeline-comparion}
\resizebox{\linewidth}{!}{

\begin{tabular}{l c c c c c}
    \toprule 
    \textbf{Object} & \textbf{Baseline} & \textbf{Depth-Only} & \textbf{RGB-Only} & \textbf{Oracle-F} & \textbf{Oracle-D} \\
    \midrule
    \multirow{2}{*}{Spray} &
    \multirow{2}{*}{81.17} &
    86.05 & 86.84 & 78.71 & 89.06 \\
    & & (+4.89) & (+5.67) & (-2.46) & (+7.89) \\
    \multirow{2}{*}{Pringles} &
    \multirow{2}{*}{61.84} &
    71.98 & 73.80 & 55.85 & 80.01 \\
    & & (+10.13) & (+11.96) & (-6.00) & (+18.17) \\
    \multirow{2}{*}{Tincase} &
    \multirow{2}{*}{63.85} &
    75.66 & 75.59 & 60.02 & 81.71 \\
    & & (+11.81) & (+11.74) & (-3.83) & (+17.86) \\
    \multirow{2}{*}{Sandwich} &
    \multirow{2}{*}{74.66} &
    75.44 & 80.03 & 71.47 & 80.85 \\
    & & (+0.78) & (+5.37) & (-3.19) & (+6.19) \\
    \multirow{2}{*}{Mouse} &
    \multirow{2}{*}{49.16} &
    55.76 & 64.02 & 47.96 & 68.91 \\
    & & (+6.60) & (+14.86) & (-1.20) & (+19.75) \\
    \multirow{2}{*}{Duck} &
    \multirow{2}{*}{74.77} &
    81.31 & 82.22 & 78.31 & 87.01 \\
    & & (+6.53) & (+7.45) & (+3.54) & (+12.24) \\
    \midrule
    \multirow{2}{*}{Overall} &
    \multirow{2}{*}{67.57} &
    74.37 & 77.08 & 65.38 & 81.26 \\
    & & (+6.79) & (+9.51) & (-2.19) & (+13.68) \\
    \bottomrule
    \end{tabular}}

}
\end{table}


\section{Evaluation on BOP metrics}
\label{appendix:bop-metrics}
\cref{tab:unseen-bop-comparison,tab:hipose-bop-comparison} report the evaluation using BOP metrics (VSD, MSSD, and MSPD) for pretrained object pose estimation models—GigaPose, SAM-6D, and FoundationPose—and HiPose.

\clearpage
\begin{sidewaystable}[htbp]
\vspace{10ex}
  \caption{\textbf{Average Recall of VSD, MSSD, and MSPD on various sensor control methods for unseen object pose estimation models: GigaPose, SAM-6D, and FoundationPose.}}
  \label{tab:unseen-bop-comparison}
  \centering
  \begin{tabular}{clccccc}
    \toprule
    \textbf{Model} & \textbf{Object} & \textbf{Baseline} & \textbf{Depth-Only} & \textbf{RGB-Only} & \textbf{Oracle-F} & \textbf{Oracle-D} \\
    \midrule

    \multirow{6}{*}{\makecell[c]{GigaPose \\ ~\cite{gigapose}}}
        & Spray
        & 72.14/84.44/84.15 & 83.86/88.39/89.76 & 82.62/92.98/93.02
        & 77.50/87.37/88.15 & 89.55/94.29/95.22 \\
        & Pringles
        & 78.62/67.95/84.62 & 86.79/81.76/94.52 & 92.54/87.05/94.81
        & 82.35/74.38/89.19 & 97.04/95.33/99.05 \\
        & Tincase
        & 91.04/96.53/99.42 & 97.56/99.69/99.96 & 93.70/97.69/99.64
        & 94.37/95.82/97.24 & 98.60/99.91/100.0 \\
        & Sandwich
        & 81.42/84.51/87.38 & 89.70/88.62/88.72 & 92.48/96.21/98.82
        & 83.07/82.10/82.77 & 98.57/99.85/100.0 \\
        & Mouse
        & 80.09/83.07/93.17 & 87.46/89.61/95.37 & 86.65/91.61/97.76
        & 77.21/77.85/86.98 & 93.53/95.95/98.88 \\
        & Duck
        & 92.79/95.40/98.74 & 97.24/98.60/100.0 & 95.21/97.95/99.49
        & 94.56/95.91/99.16 & 98.58/99.49/100.0 \\
        \cmidrule(lr){2-7}
        & Overall
        & 82.68/85.32/91.25 & 90.44/91.11/94.72 & 90.53/93.91/97.26
        & 84.84/85.57/90.58 & 95.98/97.47/98.86 \\
    \midrule
     \multirow{6}{*}{\makecell[c]{SAM-6D \\ ~\cite{sam6d}}}
      & Spray
      & 92.47/96.33/96.53 & 94.96/98.00/98.15 & 91.00/93.66/93.66
      & 92.77/96.33/96.53 & 97.43/99.22/99.17 \\
      & Pringles
      & 94.38/93.08/98.56 & 97.05/97.76/100.0 & 95.96/96.57/98.57
      & 94.32/92.43/96.81 & 99.16/99.62/100.0 \\   
      & Tincase
      & 97.21/88.08/88.90 & 96.09/91.91/92.67 & 95.87/94.22/94.84
      & 96.77/82.60/84.38 & 97.24/96.76/97.56 \\
      & Sandwich
      & 88.74/94.87/94.87 & 89.42/94.87/94.92 & 90.19/94.87/95.38
      & 88.65/94.87/94.87 & 91.32/95.44/96.31 \\
      & Mouse
      & 78.79/79.49/85.10 & 82.08/84.41/89.12 & 84.92/87.06/91.42
      & 79.04/81.56/86.87 & 88.56/89.85/93.92 \\
      & Duck
      & 93.30/94.88/96.65 & 96.01/97.12/97.86 & 97.20/98.56/99.16
      & 94.67/96.37/98.14 & 98.71/99.07/99.53 \\
      \cmidrule(lr){2-7}
      & Overall
      & 90.82/91.12/93.44 & 92.60/94.01/95.45 & 92.52/94.16/95.51
      & 91.03/90.69/92.93 & 95.40/96.66/97.75 \\

    \midrule
    \multirow{6}{*}{\makecell[c]{FoundationPose \\ ~\cite{foundationpose}}}
      & Spray
      & 89.45/99.22/99.32 & 92.85/99.76/99.85 & 94.27/100.0/100.0
      & 89.49/99.32/99.37 & 97.15/100.0/100.0 \\
      & Pringles
      & 93.98/58.29/75.29 & 96.16/77.00/86.38 & 96.41/81.10/89.52
      & 94.14/62.43/77.38 & 99.12/96.10/98.33 \\
      & Tincase
      & 97.41/43.11/47.51 & 97.64/62.62/64.98 & 97.72/77.73/80.58
      & 97.00/45.33/47.69 & 97.75/91.47/93.33 \\
      & Sandwich
      & 94.49/95.08/96.31 & 97.26/97.59/98.97 & 95.75/96.26/98.26
      & 95.87/97.38/97.85 & 97.96/98.41/99.49 \\
      & Mouse
      & 88.39/90.59/94.73 & 91.24/92.59/97.66 & 90.72/91.95/96.68
      & 89.02/91.56/96.54 & 93.66/94.15/98.44 \\
      & Duck
      & 98.63/99.42/100.0 & 99.41/99.63/100.0 & 99.85/99.95/100.0
      & 98.64/99.57/100.0 & 99.97/100.0/100.0 \\
      \cmidrule(lr){2-7}
      & Overall
      & 93.72/80.95/85.53 & 95.76/88.20/91.31 & 95.79/91.16/94.17
      & 94.03/82.60/86.47 & 97.60/96.69/98.27 \\
    \bottomrule
  \end{tabular}
\end{sidewaystable}

\begin{sidewaystable}[htbp]
\vspace{5ex}
  \caption{\textbf{Average Recall of VSD, MSSD, and MSPD on various sensor control methods for HiPose.}}
  \label{tab:hipose-bop-comparison}
  \centering
  \begin{tabular}{lccccc}
    \toprule
    \textbf{Object} & \textbf{Baseline} & \textbf{Depth-Only} & \textbf{RGB-Only} & \textbf{Oracle-F} & \textbf{Oracle-D} \\
    \midrule
    Spray
    & 94.63/99.46/99.56 & 96.76/99.56/99.61 & 97.51/99.90/99.95
    & 78.06/82.39/82.39 & 99.05/99.95/100.0 \\
    Pringles
    & 98.10/96.71/100.0 & 98.94/97.90/100.0 & 99.67/99.33/100.0
    & 85.22/84.67/88.38 & 99.82/99.76/100.0 \\
    Tincase
    & 89.26/88.27/90.67 & 91.65/90.62/92.76 & 97.26/97.33/98.44
    & 78.14/78.22/78.98 & 97.38/97.38/98.84 \\
    Sandwich
    & 92.17/92.77/93.28 & 94.62/95.23/95.54 & 95.49/95.74/96.82
    & 76.65/77.69/78.21 & 97.06/97.38/98.31 \\
    Mouse 
    & 79.08/82.98/88.39 & 82.08/84.98/91.32 & 86.98/89.46/94.98
    & 65.81/68.98/73.90 & 90.93/92.73/97.46 \\
    Duck 
    & 80.92/81.86/83.07 & 82.99/83.86/84.84 & 98.71/99.40/99.77
    & 77.23/78.28/79.35 & 99.31/99.53/99.86 \\
    \midrule
    Overall 
    & 89.03/90.34/92.49 & 91.17/92.03/94.01 & 95.94/96.86/98.33
    & 76.85/78.37/80.20 & 97.26/97.79/99.08 \\
    
    \bottomrule    
  \end{tabular}%
\end{sidewaystable}

\clearpage
\section{RGB Augmentation Settings} 
\label{appendix:augmentation}

\begin{figure}[!t]
\vspace{-30ex}
  \centering{
  \includegraphics[width=0.5\textwidth]{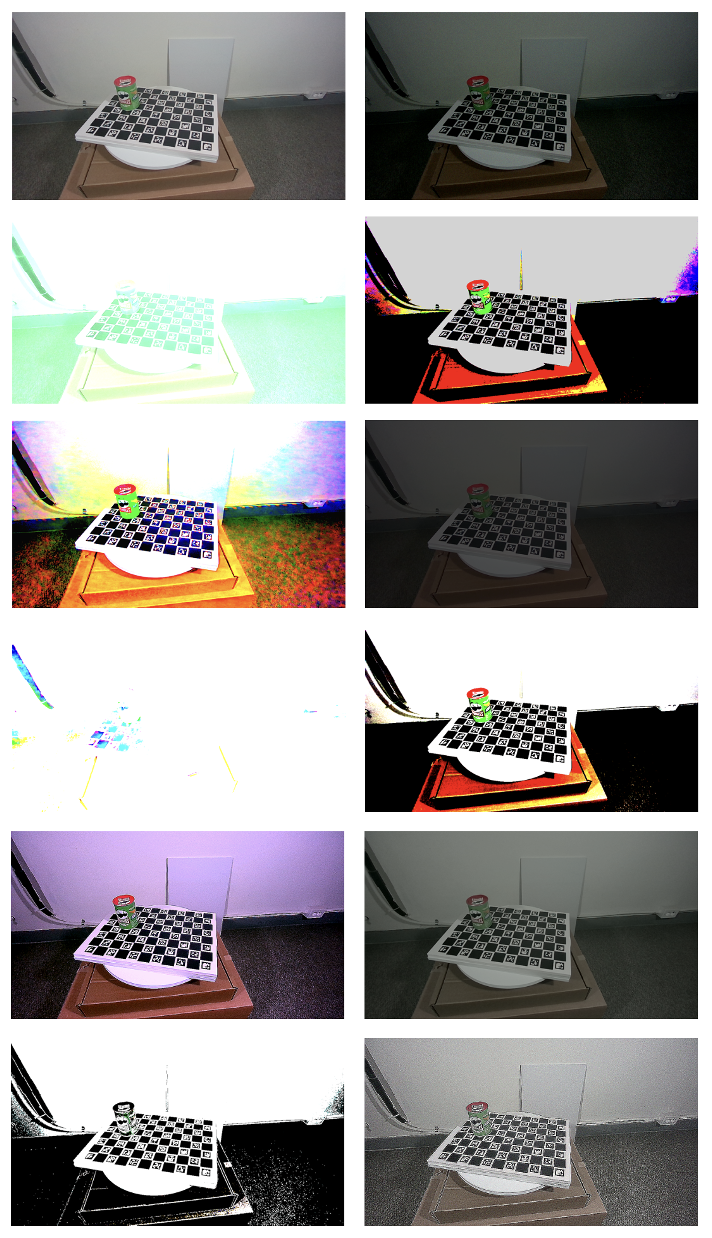}
    \caption{\textbf{Examples of RGB augmentations applied during training.} The top-left image is the original, and the remaining images are randomly augmented samples generated using the settings summarized in~\cref{tab:aug_list}.}
  \label{fig:augmentation_examples}
  }
\end{figure}

To improve generalization under varying environmental conditions, we adopted an RGB augmentation pipeline based on that used in GDRNPP. The augmentations were implemented using the \texttt{imgaug} library and applied only to the RGB input during training with a probability of 0.8. An overview of these augmentations is illustrated in~\cref{fig:augmentation_examples}.

The augmentations listed in~\cref{tab:aug_list} were applied in random order, with each operation activated independently with a fixed probability.

\begin{table}[h]
\centering{
\caption{\textbf{Augmentation operations and parameters.}}
\begin{tabular}{l l l}
\toprule
\makecell[c]{\textbf{Augmentation} \\ \textbf{Type}} &
\makecell[c]{\textbf{Activation} \\ \textbf{Probability}} &
\makecell[c]{\textbf{Parameters} \\ \textbf{/ Range}} \\
\midrule
\makecell[l]{Gaussian \\ Blur} & 0.4 & $\sigma \in [0.0, 3.0]$ \\
\makecell[l]{Sharpness \\ Enhancement} & 0.3 & factor $\in [0.0, 50.0]$ \\
\makecell[l]{Contrast \\ Enhancement} & 0.3 & factor $\in [0.2, 50.0]$ \\
\makecell[l]{Brightness \\ Enhancement} & 0.5 & factor $\in [0.1, 6.0]$ \\
\makecell[l]{Color \\ Enhancement} & 0.3 & factor $\in [0.0, 20.0]$ \\
Additive Intensity & 0.5 & \makecell[l]{value $\in [-25, 25]$, \\ per-channel = 0.3}  \\
Invert Pixels & 0.3 & \makecell[l]{probability = 0.2, \\ per-channel = } \\
Multiply Intensity & 0.5 & \makecell[l]{factor $\in [0.6, 1.4]$, \\ per-channel = 0.5} \\
\makecell[l]{Additive \\ Gaussian Noise} & 0.1 & \makecell[l]{scale = 10, \\ per-channel} \\
\makecell[l]{Linear Contrast} & 0.5 & \makecell[l]{factor $\in [0.5, 2.2]$, \\ per-channel = 0.3} \\
\makecell[l]{Grayscale \\ Conversion} & 0.5 & $\alpha \in [0.0, 1.0]$ \\
\bottomrule
\end{tabular}
\label{tab:aug_list}
}
\end{table}

\clearpage
\section{Evaluation on Markerless Test Scenes}
\label{appendix:markerless}
\begin{figure}[h]
  \centering{
  \includegraphics[width=0.345\textwidth]{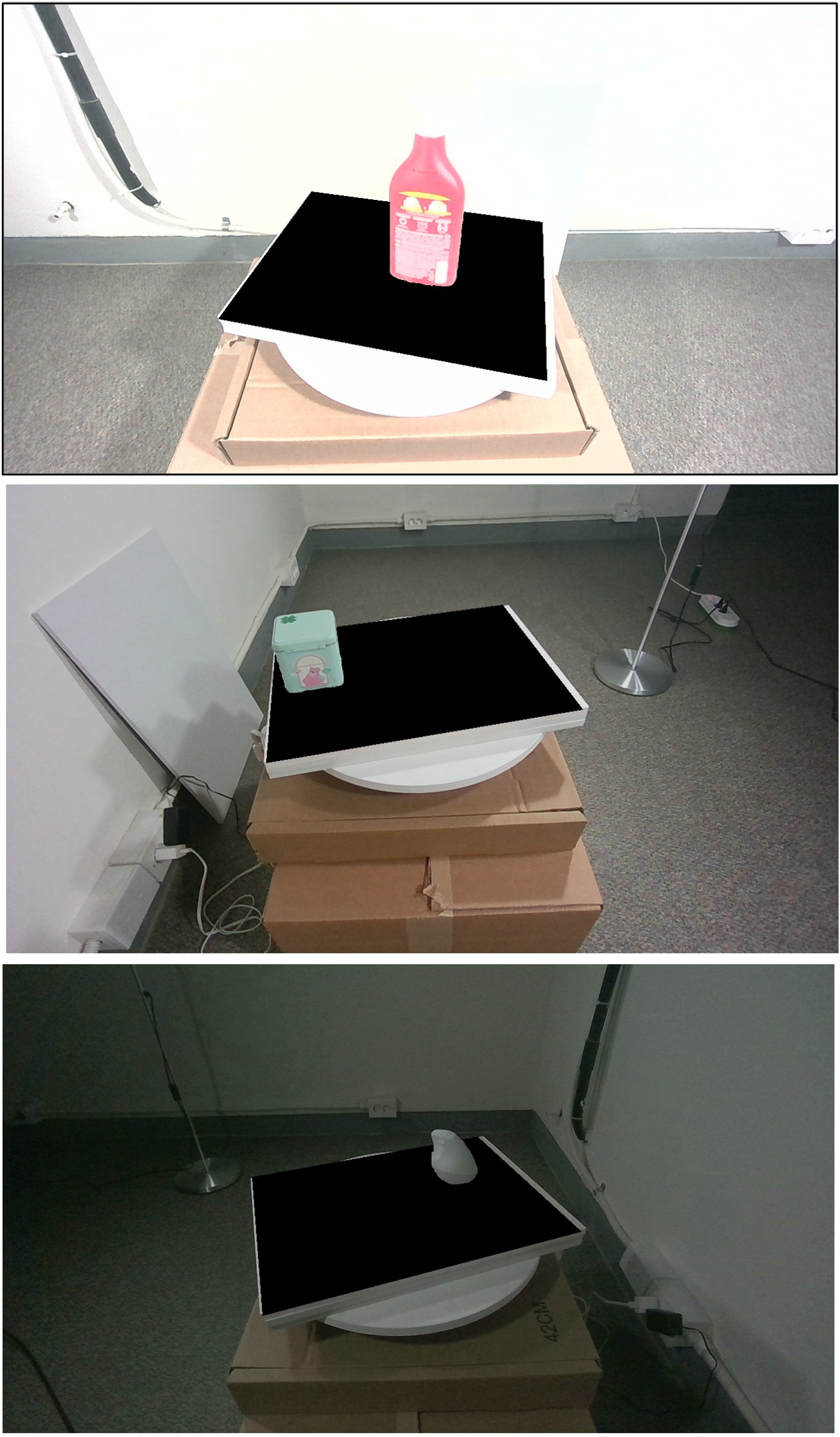}
  \caption{
    \textbf {Sample of marker-removed test images.}
    }
  \label{fig:appendix_markerless}
  }
\end{figure}

Since visible markers in the scene can introduce model bias~\cite{markerBias}, we modified Test-Var split images (see~\cref{fig:appendix_markerless}) and evaluated them using (i) the same model weights as in~\cref{sec:instance-level} (\cref{tab:markerless-result}) and (ii) models trained on the PBR + marker-removed Train-Def setting (\cref{tab:supervised-markerless-result}).

In \cref{tab:markerless-result}, the RGB-D based model (i.e., HiPose~\cite{hipose}) maintains performance comparable to \cref{tab:pose-comparison}, whereas the RGB-based models (i.e., ZebraPose~\cite{zebrapose} and GDRNPP~\cite{gdrnpp}) show substantial object-dependent variation.
Both RGB models retain similar performance for Spray compared to \cref{tab:pose-comparison}, but exhibit a dramatic drop for Mouse. Meanwhile, \cref{tab:supervised-markerless-result} shows overall performance comparable to \cref{tab:pose-comparison}.
These results suggest that Spray, which has rich RGB features, can be learned based on its distinctive texture. In contrast, Mouse, which has an ambiguous shape and pale coloration, lacks discriminative features, making the models more susceptible to domain shift.
Despite these discrepancies, sensor optimization consistently adapts to environmental changes more effectively than existing generalization techniques, even without additional training.

\begin{table*}[ht]
\vspace{-5ex}
  \caption{
    \textbf{AR@5 comparison of sensor control methods on marker-removed \ours with ZebraPose, GDRNPP, and HiPose.}}
    
  \label{tab:markerless-result}
  \centering
  \resizebox{\linewidth}{!}{
  \begin{tabular}{ll cccc cccc cccc}
    \toprule
    \multicolumn{2}{c}{} 
    & \multicolumn{4}{c}{\textbf{ZebraPose~\cite{zebrapose}}}
    & \multicolumn{4}{c}{\textbf{GDRNPP~\cite{gdrnpp}}}
    & \multicolumn{4}{c}{\textbf{HiPose~\cite{hipose}}} \\
    \cmidrule(lr){3-6} \cmidrule(lr){7-10} \cmidrule(lr){11-14}
    \multicolumn{2}{c}{} 
    & \textbf{Test-Def} & \multicolumn{3}{c}{\textbf{Test-Var}}
    & \textbf{Test-Def} & \multicolumn{3}{c}{\textbf{Test-Var}}
    & \textbf{Test-Def} & \multicolumn{3}{c}{\textbf{Test-Var}} \\
    \cmidrule(lr){3-6} \cmidrule(lr){7-10} \cmidrule(lr){11-14}
    \textbf{Object} & \textbf{Train} 
      & Auto & Auto & Rand & Oracle 
      & Auto & Auto & Rand & Oracle 
      & Auto & Auto & Rand & Oracle \\
    \midrule

    \multirow{3}{*}{Spray}
      & PBR + Train-Def
        & 96.59 & 95.12 & 64.33 & 100
        & 100 & 96.59 & 52.49 & 99.51
        & 100 & 98.73 & 59.98 & 100  \\
      & PBR + Train-Def w/ Aug
        & 93.66 & 93.17 & 70.71 & 100
        & 95.12 & 94.63 & 71.63 & 100
        & 100 & 100 & 91.66 & 100 \\
      & PBR + Train-Var
        & 98.05 & 97.07 & 78.91 & 100
        & 100 & 100 & 81.89 & 100
        & 100 & 100 & 100 & 100 \\

    \midrule
    \multirow{3}{*}{Tincase}
      & PBR + Train-Def
        & 94.22 & 86.22 & 55.45 & 99.11
        & 64.44 & 58.66 & 35.07 & 84.89
        & 97.78 & 84.62 & 56.33 & 97.78 \\
      & PBR + Train-Def w/ Aug
        & 76.44 & 72.89 & 52.62 & 87.11
        & 40 & 38.22 & 26.69 & 61.33
        & 97.78 & 96.89 & 88.40 & 97.78 \\
      & PBR + Train-Var
        & 85.33 & 81.33 & 59.79 & 96.00
        & 51.11 & 48.44 & 34.40 & 71.56
        & 97.78 & 97.69 & 97.73 & 97.78 \\

    \midrule
    \multirow{3}{*}{Mouse}
      & PBR + Train-Def
        & 48.78 & 43.41 & 30.63 & 84.88
        & 4.88 & 3.90 & 2.98 & 29.27
        & 90.24 & 67.42 & 47.17 & 88.78 \\
      & PBR + Train-Def w/ Aug
        & 26.83 & 24.39 & 18.83 & 57.56
        & 12.2 & 12.17 & 11.66 & 45.37
        & 90.24 & 88.49 & 84.56 & 90.24 \\
      & PBR + Train-Var
        & 9.76 & 8.29 & 5.07 & 25.37
        & 2.44 & 1.95 & 3.39 & 19.02
        & 90.24 & 90.15 & 87.68 & 90.73 \\
    \bottomrule
  \end{tabular}}
  \vspace{-1ex}
\end{table*}

\begin{table*}[ht]
  \caption{
    \textbf{AR@5 comparison of sensor control methods on marker-removed \ours with ZebraPose, GDRNPP, and HiPose trained on PBR + marker-removed Train-Def setting.}
    }
  \label{tab:supervised-markerless-result}
  \centering
  \resizebox{\linewidth}{!}{
  \begin{tabular}{ll cccc cccc cccc}
    \toprule
    \multicolumn{2}{c}{} 
    & \multicolumn{4}{c}{\textbf{ZebraPose~\cite{zebrapose}}}
    & \multicolumn{4}{c}{\textbf{GDRNPP~\cite{gdrnpp}}}
    & \multicolumn{4}{c}{\textbf{HiPose~\cite{hipose}}} \\
    \cmidrule(lr){3-6} \cmidrule(lr){7-10} \cmidrule(lr){11-14}
    \multicolumn{2}{c}{} 
    & \textbf{Test-Def} & \multicolumn{3}{c}{\textbf{Test-Var}}
    & \textbf{Test-Def} & \multicolumn{3}{c}{\textbf{Test-Var}}
    & \textbf{Test-Def} & \multicolumn{3}{c}{\textbf{Test-Var}} \\
    \cmidrule(lr){3-6} \cmidrule(lr){7-10} \cmidrule(lr){11-14}
    \textbf{Object} & \textbf{Train} 
      & Auto & Auto & Rand & Oracle 
      & Auto & Auto & Rand & Oracle 
      & Auto & Auto & Rand & Oracle \\
    \midrule

    \multirow{3}{*}{Spray}
      & PBR + Train-Def
        & 100.0 & 97.56 & 55.95 & 99.51
        & 100.0 & 95.12 & 50.76 & 98.54
        & 100.0 & 97.07 & 56.49 & 100.0 \\
      & PBR + Train-Def w/ Aug
        & 97.56 & 89.78 & 66.95 & 100.00
        & 97.56 & 97.07 & 76.10 & 99.51
        & 100.0 & 100.0 & 98.71 & 100.0 \\
      & PBR + Train-Var
        & 95.12 & 90.73 & 73.85 & 97.56
        & 95.12 & 95.12 & 87.61 & 95.12
        & 100.0 & 100.0 & 99.73 & 100.0  \\

    \midrule
    \multirow{3}{*}{Tincase}
      & PBR + Train-Def
        & 97.78 & 96.10 & 50.89 & 98.67
        & 93.33 & 82.22 & 47.60 & 94.67
        & 97.33 & 82.49 & 53.82 & 96.44 \\
      & PBR + Train-Def w/ Aug
        & 97.78 & 92.00 & 64.82 & 100.0
        & 91.11 & 87.55 & 66.51 & 95.56
        & 97.78 & 95.82 & 77.20 & 97.78 \\
      & PBR + Train-Var
        & 95.56 & 95.56 & 82.09 & 97.78
        & 100.0 & 99.56 & 93.16 & 100.0
        & 97.78 & 97.78 & 97.73 & 97.78  \\
    \midrule
    \multirow{3}{*}{Mouse}
      & PBR + Train-Def
        & 75.61 & 65.85 & 27.59 & 92.68
        & 90.24 & 75.12 & 32.07 & 88.29 
        & 85.85 & 70.54 & 45.61 & 89.27 \\

      & PBR + Train-Def w/ Aug
        & 51.22 & 56.59 & 46.46 & 94.63
        & 73.17 & 68.78 & 47.93 & 88.78
        &  90.24 & 89.56 & 73.17 & 90.24 \\

      & PBR + Train-Var
        & 58.54 & 56.59 & 48.37 & 79.51
        & 73.17 & 73.66 & 69.02 & 85.85
        & 90.24 & 90.05 & 89.17 & 90.73 \\

    \bottomrule
  \end{tabular}}
  \vspace{-1ex}
\end{table*}

\section{Factor-wise Performance across Brightness Levels}
\label{appendix: factorwise-performance}
\begin{figure*}[h]
  \centering{
  \includegraphics[width=\textwidth]{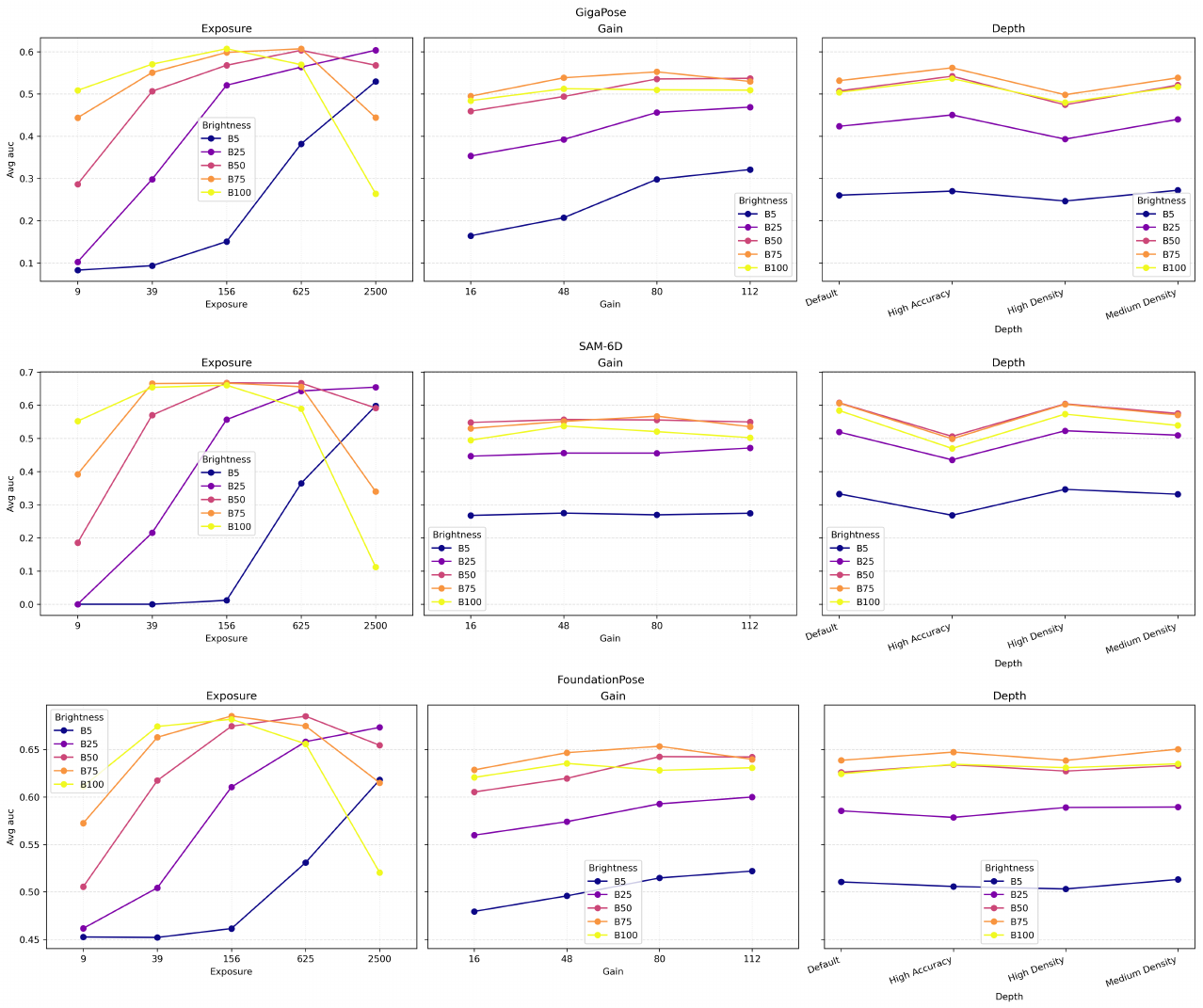}
  \caption{
    \textbf {Average AUC by brightness level across exposure, gain, and depth capture modes for pretrained models:} GigaPose, SAM-6D, and FoundationPose.}
  \label{figs:factorwise_pretrained}
  }
\end{figure*}

\begin{figure*}[h]
  \centering{
  \includegraphics[width=\textwidth]{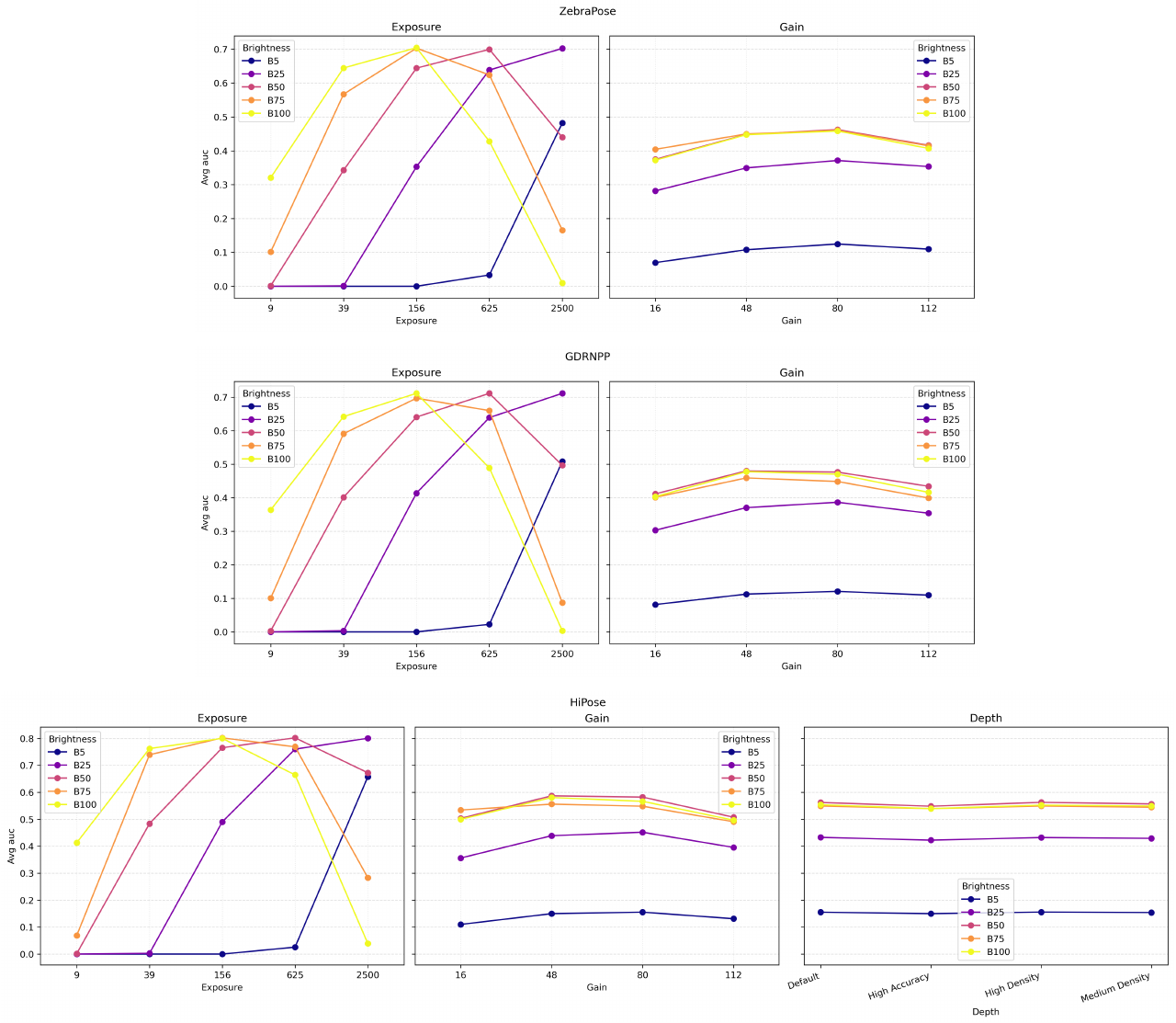}
  \caption{
    \textbf {Average AUC by brightness level across exposure, gain, and depth capture modes for supervised models:} ZebraPose, GDRNPP, and HiPose.}
  \label{figs:factorwise_supervised}
  }
\end{figure*}

\cref{figs:factorwise_pretrained},~\cref{figs:factorwise_supervised} show how model performance varies with sensor settings (exposure, gain, and depth) at each brightness level for pretrained and supervised models. Performance varied most clearly with exposure across brightness conditions. In darker scenes (B5 and B25), AUC generally increased with exposure. In brighter scenes, the best-performing exposure shifted lower, and higher exposure sometimes reduced performance. AUC changes across gain and depth capture modes were comparatively small, and their patterns differed across models.

\section{Qualitative Analysis}
\subsection{Qualitative Result on Pretrained, Generalizable Pose Estimation Models}
\label{appendix: qualitative-unseen}
\begin{figure*}[h]
  \centering{
  \includegraphics[width=\textwidth]{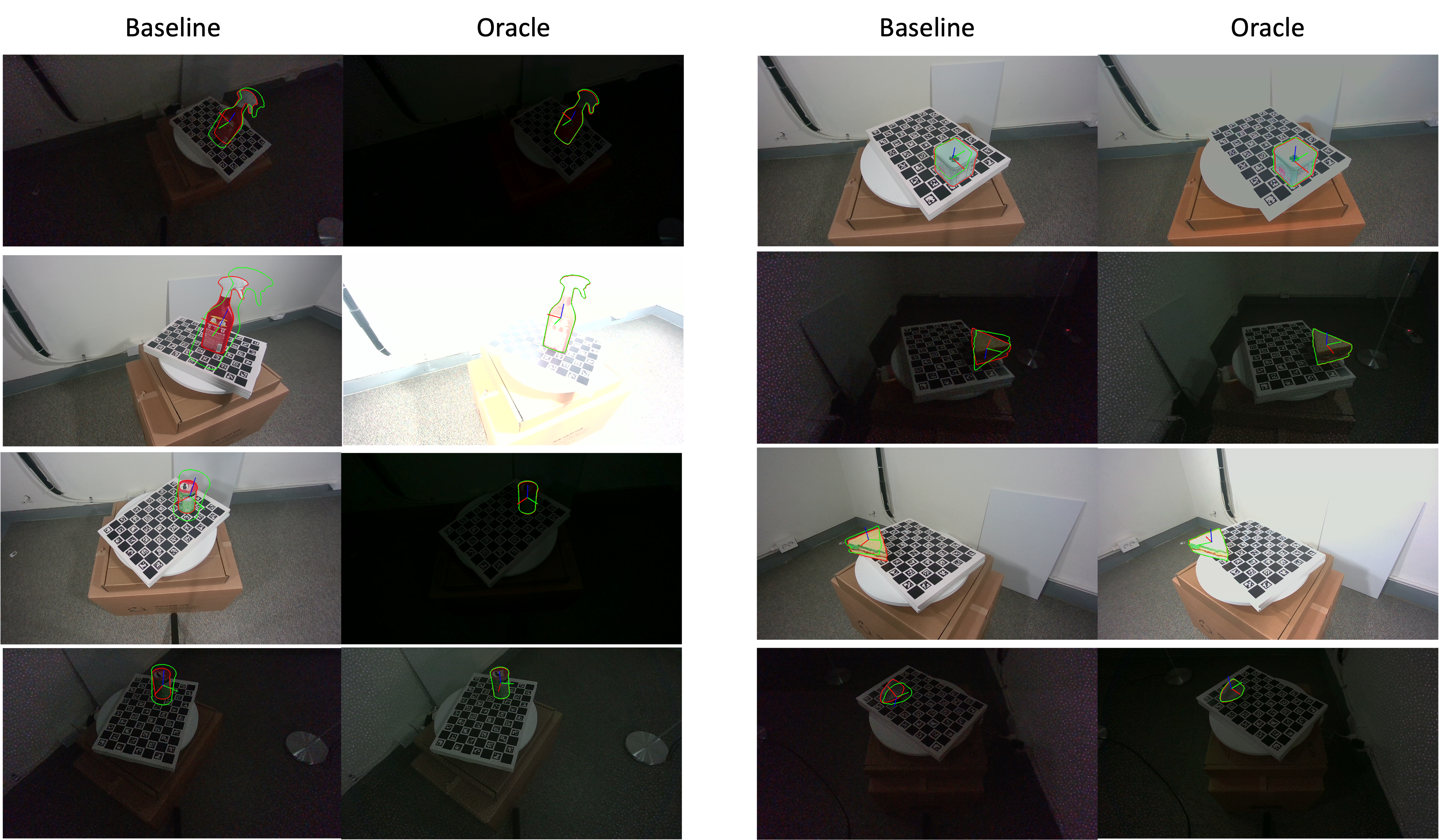}
  \caption{
    \textbf {Comparison of predictions under Baseline and Oracle for GigaPose.} Visualized object pose on RGB images: ground truth pose in red, predicted pose in green.
    }
  \label{fig:appendix_qualitative_gigapose}
  }
\end{figure*}

\begin{figure*}[h]
  \centering{
  \includegraphics[width=\textwidth]{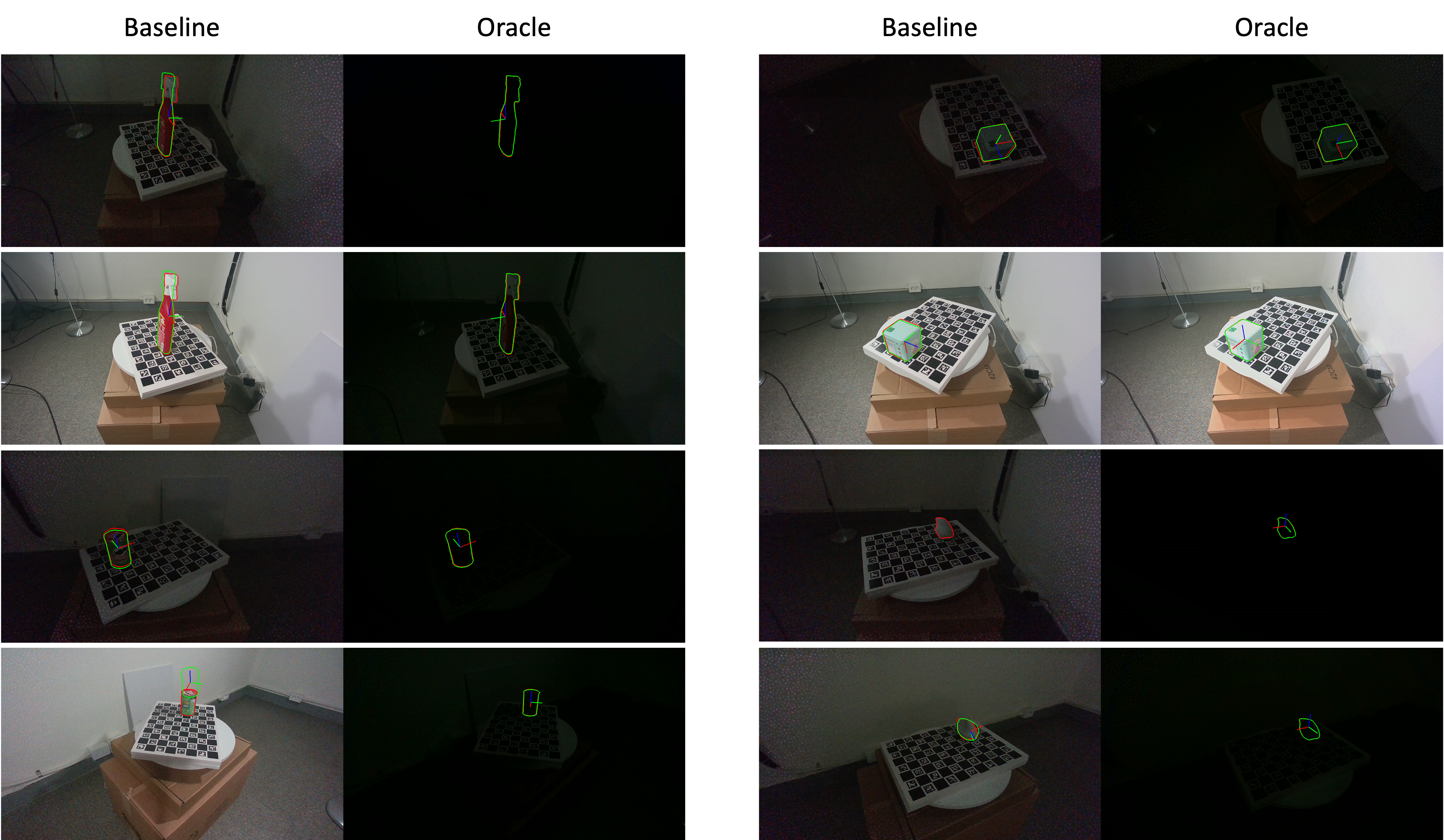}
  \caption{
    \textbf {Comparison of predictions under Baseline and Oracle for SAM-6D.} Visualized object pose on RGB images: ground truth pose in red, predicted pose in green.
    }
  \label{fig:appendix_qualitative_sam6d}
  }
\end{figure*}

\begin{figure*}[h]
  \centering{
  \vspace{-4ex}
  \includegraphics[width=\textwidth]{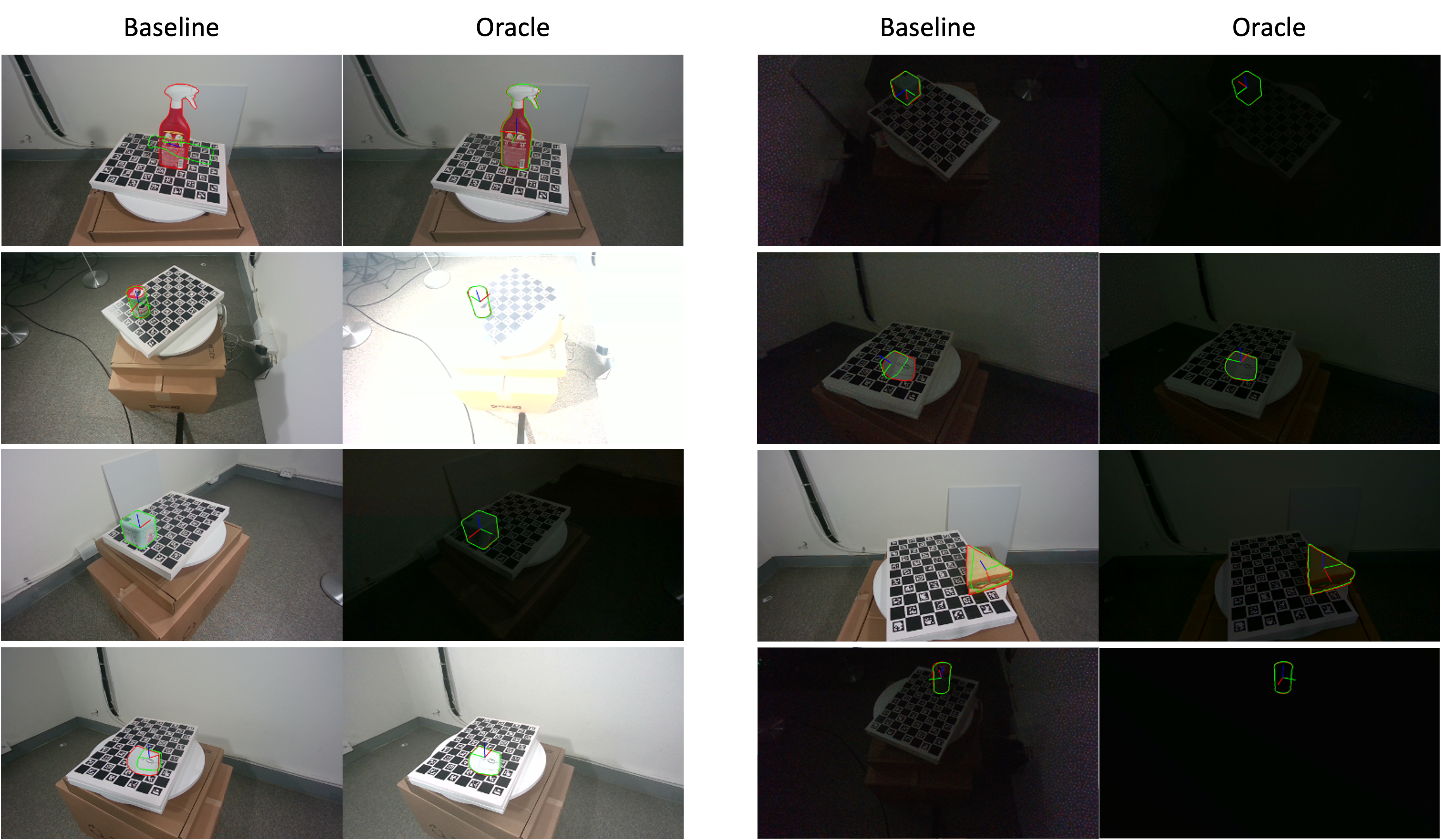}
  \caption{
    \textbf {Comparison of predictions under Baseline and Oracle for FoundationPose.} Visualized object pose on RGB images: ground truth pose in red, predicted pose in green.
    }
  \label{fig:appendix_qualitative_foundationpose}
  }
\end{figure*}

\cref{fig:appendix_qualitative_gigapose},~\cref{fig:appendix_qualitative_sam6d},~\cref{fig:appendix_qualitative_foundationpose} visualize GT and predicted poses on Baseline and Oracle sensor configuration for GigaPose, SAM-6D and FoundationPose.

\subsection{Qualitative Result on Instance-Level Pose Estimation Model}
\label{appendix:qualitative-instance}
\begin{figure*}[h]
\vspace{-14ex}
  \centering{
  \includegraphics[width=\textwidth]{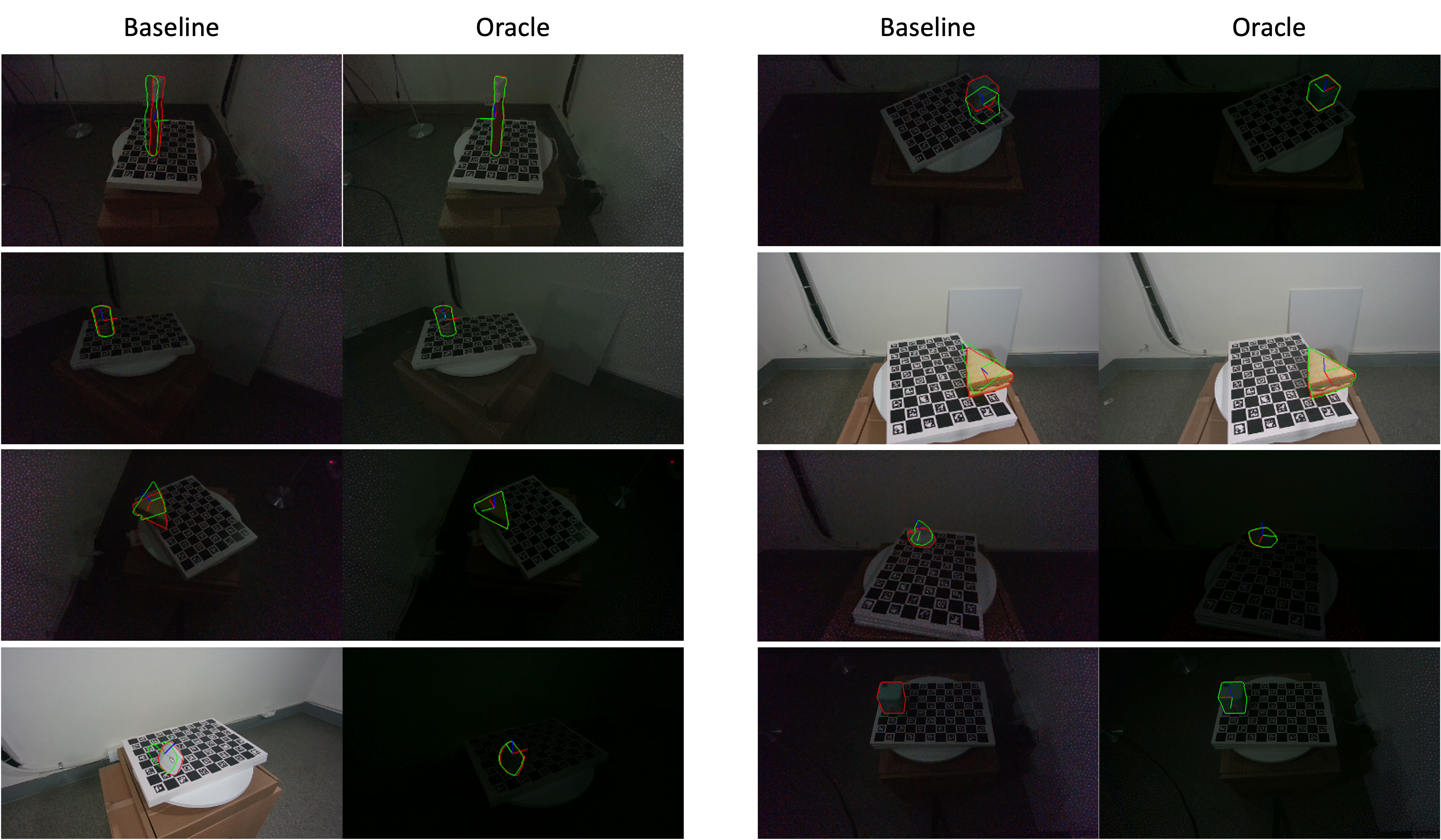}
  \caption{
    \textbf {Comparison of predictions under Baseline and Oracle for HiPose.} Visualized object pose on RGB images: ground truth pose in red, predicted pose in green.
    }
  \label{fig:appendix_qualitative_hipose}
  }
\end{figure*}

\cref{fig:appendix_qualitative_hipose} visualizes GT and predicted poses on Baseline and Oracle sensor configuration for HiPose.

\section{Evaluation on RGB-D Based Models Under Different Environments}
\cref{tab:unseen-environments-hipose},~\cref{tab:unseen-environments-gigapose},~\cref{tab:unseen-environments-sam6d},~\cref{tab:unseen-environments-foundationpose} shows performance of RGB-D models---HiPose, GigaPose, SAM-6D, and FoundationPose---under different illumination conditions.

\begin{table*}[ht]
\centering

\caption{\textbf{AUC@[0:0.1] performance under diverse environments of various sensor control methods for HiPose trained on PBR + Train-Def.}}
\label{tab:unseen-environments-hipose}
\resizebox{1.0\linewidth}{!}{
\begin{tabular}{clccccc}
    \toprule
    \textbf{Env} & \textbf{Object} & \textbf{Baseline} & \textbf{Depth-Only} & \textbf{RGB-Only} & \textbf{Oracle-F} & \textbf{Oracle-D} \\
    \midrule
    \multirow{6}{*}{\makecell[c]{B5}}
      & Spray
      & 83.17 & 85.07 (+1.90) & 87.10 (+3.93) & 86.07 (+2.90) & 88.63 (+5.46) \\
      & Pringles
      & 72.40 & 75.45 (+3.05) & 78.60 (+6.19) & 77.90 (+5.50) & 81.12 (+8.71) \\
      & Tincase 
      & 32.96 & 43.00 (+10.04) & 78.80 (+45.84) & 76.69 (+43.73) & 81.31 (+48.36) \\
      & Sandwich
      & 66.85 & 69.85 (+3.00) & 80.72 (+13.87) & 78.82 (+11.97) & 83.18 (+16.33) \\
      & Mouse
      & 35.29 & 39.88 (+4.59) & 64.85 (+29.56) & 62.93 (+27.63) & 69.88 (+34.59) \\
      & Duck 
      & 5.93 & 12.05 (+6.12) & 79.16 (+73.23) & 76.14 (+70.21) & 81.44 (+75.51) \\
      \cmidrule(lr){2-7}
      & Overall
      & 49.43 & 54.22 (+4.78) & 78.20 (+28.77) & 76.43 (+26.99) & 80.93 (+31.49) \\

    \midrule
    \multirow{6}{*}{\makecell[c]{B25}}
      & Spray
      & 87.56 & 88.71 (+1.15) & 88.61 (+1.05) & 87.59 (+0.02) & 89.63 (+2.07) \\
      & Pringles
      & 74.12 & 77.17 (+3.05) & 79.14 (+5.02) & 77.31 (+3.19) & 81.74 (+7.62) \\
      & Tincase
      & 85.20 & 86.33 (+1.13) & 86.91 (+1.71) & 85.13 (-0.07) & 87.89 (+2.69) \\
      & Sandwich
      & 80.54 & 83.51 (+2.97) & 83.31 (+2.77) & 82.97 (+2.44) & 86.51 (+5.97) \\
      & Mouse
      & 70.41 & 73.17 (+2.76) & 74.32 (+3.90) & 71.54 (+1.12) & 77.29 (+6.88) \\
      & Duck
      & 85.88 & 86.93 (+1.05) & 88.47 (+2.58) & 87.16 (+1.28) & 90.02 (+4.14) \\
      \cmidrule(lr){2-7}
      & Overall
      & 80.62 & 82.64 (+2.02) & 83.46 (+2.84) & 81.95 (+1.33) & 85.51 (+4.90) \\

    \midrule
    \multirow{6}{*}{\makecell[c]{B50}}
      & Spray    
      & 87.93 & 88.78 (+0.85) & 89.32 (+1.39) & 88.07 (+0.15) & 90.07 (+2.15) \\
      & Pringles
      & 74.24 & 76.57 (+2.33) & 79.69 (+5.45) & 76.74 (+2.50) & 81.62 (+7.38) \\
      & Tincase
      & 85.11 & 86.69 (+1.58) & 86.91 (+1.80) & 85.29 (+0.18) & 88.31 (+3.20) \\
      & Sandwich 
      & 81.38 & 84.38 (+3.00) & 85.72 (+4.33) & 84.03 (+2.64) & 87.23 (+5.85) \\
      & Mouse
      & 71.93 & 73.17 (+1.24) & 74.51 (+2.59) & 71.63 (-0.29) & 77.44 (+5.51) \\
      & Duck
      & 86.77 & 88.51 (+1.74) & 88.63 (+1.86) & 87.51 (+0.74) & 90.49 (+3.72) \\
      \cmidrule(lr){2-7}
      & Overall
      & 81.23 & 83.02 (+1.79) & 84.13 (+2.90) & 82.21 (+0.99) & 85.86 (+4.63)  \\

    \midrule
    \multirow{6}{*}{\makecell[c]{B75}}
      & Spray
      & 88.02 & 89.02 (+1.00) & 89.51 (+1.49) & 88.02 (+0.00) & 90.61 (+2.59) \\
      & Pringles
      & 73.31 & 76.02 (+2.71) & 79.12 (+5.81) & 76.83 (+3.52) & 81.86 (+8.55) \\
      & Tincase
      & 85.27 & 86.49 (+1.22) & 86.80 (+1.53) & 85.33 (+0.07) & 88.13 (+2.87) \\
      & Sandwich
      & 82.10 & 85.13 (+3.03) & 84.41 (+2.31) & 84.38 (+2.28) & 87.10 (+5.00) \\
      & Mouse
      & 71.73 & 73.27 (+1.54) & 74.17 (+2.44) & 71.24 (-0.49) & 76.07 (+4.34) \\
      & Duck
      & 86.93 & 88.70 (+1.77) & 88.67 (+1.74) & 86.35 (-0.58) & 89.79 (+2.86) \\
      \cmidrule(lr){2-7}
      & Overall
      & 81.23 & 83.11 (+1.88) & 83.78 (+2.55) & 82.03 (+0.80) & 85.59 (+4.37) \\

    \midrule
    \multirow{6}{*}{\makecell[c]{B100}}
      & Spray    & 87.93 & 89.27 (+1.34) & 89.39 (+1.46) & 87.95 (+0.02) & 90.41 (+2.49) \\
      & Pringles & 72.90 & 75.69 (+2.79) & 78.86 (+5.95) & 76.31 (+3.40) & 81.40 (+8.50) \\
      & Tincase  & 85.38 & 86.87 (+1.49) & 87.09 (+1.71) & 85.40 (+0.02) & 88.53 (+3.16) \\
      & Sandwich & 82.51 & 85.72 (+3.21) & 86.15 (+3.64) & 84.31 (+1.80) & 88.10 (+5.59) \\
      & Mouse    & 70.24 & 72.59 (+2.34) & 73.07 (+2.83) & 70.24 (+0.00) & 75.39 (+5.15) \\
      & Duck
      & 85.86 & 88.53 (+2.67) & 88.44 (+2.58) & 87.12 (+1.26) & 90.56 (+4.70) \\
      \cmidrule(lr){2-7}
      & Overall
      & 80.80 & 83.11 (+2.31) & 83.83 (+3.03) & 81.89 (+1.08) & 85.73 (+4.93) \\
    \bottomrule
  \end{tabular}}
\end{table*}

\begin{table*}[htbp]
\centering
\caption{\textbf{AUC@[0:0.1] performance under diverse environments of various sensor control methods for GigaPose.}}
\label{tab:unseen-environments-gigapose}
\resizebox{1.0\linewidth}{!}{
  \begin{tabular}{clccccc}
    \toprule
    \textbf{Env} & \textbf{Object} & \textbf{Baseline} & \textbf{Depth-Only} & \textbf{RGB-Only} & \textbf{Oracle-F} & \textbf{Oracle-D} \\
    \midrule

\multirow{6}{*}{\makecell[c]{B5}}
      & Spray
      & 65.02 & 71.00 (+5.98) & 70.22 (+5.20) & 66.85 (+1.83) & 75.54 (+10.51) \\
      & Pringles
      & 22.95 & 42.05 (+19.10) & 42.93 (+19.98) & 26.79 (+3.83) & 64.60 (+41.64) \\
      & Tincase
      & 65.98 & 78.67 (+12.69) & 71.69 (+5.71) & 74.07 (+8.09) & 80.27 (+14.29) \\
      & Sandwich
      & 59.97 & 69.74 (+9.77) & 68.77 (+8.79) & 59.97 (+0.00) & 80.31 (+20.33) \\
      & Mouse
      & 44.05 & 55.17 (+11.12) & 62.73 (+18.68) & 56.32 (+12.27) & 72.78 (+28.73) \\
      & Duck
      & 72.88 & 77.56 (+4.67) & 75.67 (+2.79) & 72.44 (-0.44) & 80.30 (+7.42) \\
      \cmidrule(lr){2-7}
      & Overall
      & 55.14 & 65.70 (+10.55) & 65.34 (+10.19) & 59.41 (+4.26) & 75.63 (+20.49) \\

    \midrule
    \multirow{6}{*}{\makecell[c]{B25}}
      & Spray
      & 68.37 & 72.90 (+4.54) & 77.44 (+9.07) & 70.78 (+2.41) & 80.54 (+12.17) \\
      & Pringles
      & 19.12 & 47.64 (+28.52) & 46.29 (+27.17) & 41.55 (+22.43) & 69.64 (+50.52) \\
      & Tincase
      & 72.69 & 81.91 (+9.22) & 77.22 (+4.53) & 80.36 (+7.67) & 83.58 (+10.89) \\
      & Sandwich
      & 67.87 & 74.33 (+6.46) & 77.54 (+9.67) & 76.10 (+8.23) & 84.95 (+17.08) \\
      & Mouse
      & 56.76 & 66.80 (+10.05) & 66.78 (+10.02) & 61.78 (+5.02) & 74.10 (+17.34) \\
      & Duck
      & 72.05 & 78.60 (+6.56) & 77.53 (+5.49) & 74.84 (+2.79) & 82.70 (+10.65) \\
      \cmidrule(lr){2-7}
      & Overall
      & 59.47 & 70.37 (+10.89) & 70.47 (+10.99) & 67.57 (+8.09) & 79.25 (+19.78) \\

    \midrule
    \multirow{6}{*}{\makecell[c]{B50}}
      & Spray
      & 64.44 & 73.24 (+8.80) & 76.71 (+12.27) & 72.22 (+7.78) & 80.83 (+16.39) \\
      & Pringles
      & 23.93 & 44.26 (+20.33) & 43.50 (+19.57) & 35.81 (+11.88) & 69.76 (+45.83) \\
      & Tincase
      & 77.40 & 81.96 (+4.56) & 78.80 (+1.40) & 80.42 (+3.02) & 84.09 (+6.69) \\
      & Sandwich
      & 62.13 & 70.08 (+7.95) & 77.49 (+15.36) & 77.77 (+15.64) & 84.62 (+22.49) \\
      & Mouse
      & 58.12 & 68.98 (+10.85) & 67.95 (+9.83) & 63.88 (+5.76) & 77.73 (+19.61) \\
      & Duck
      & 71.79 & 79.84 (+8.05) & 78.93 (+7.14) & 75.84 (+4.05) & 82.95 (+11.16) \\
      \cmidrule(lr){2-7}
      & Overall
      & 59.63 & 69.73 (+10.09) & 70.56 (+10.93) & 67.66 (+8.02) & 80.00 (+20.36) \\

    \midrule
    \multirow{6}{*}{\makecell[c]{B75}}
      & Spray
      & 60.85 & 69.68 (+8.83) & 77.07 (+16.22) & 73.32 (+12.46) & 80.76 (+19.90) \\
      & Pringles
      & 23.55 & 45.19 (+21.64) & 49.88 (+26.33) & 35.98 (+12.43) & 71.79 (+48.24) \\
      & Tincase
      & 77.49 & 81.53 (+4.04) & 80.71 (+3.22) & 80.51 (+3.02) & 84.07 (+6.58) \\
      & Sandwich
      & 68.62 & 75.33 (+6.72) & 77.67 (+9.05) & 78.69 (+10.08) & 85.21 (+16.59) \\
      & Mouse
      & 56.02 & 68.12 (+12.10) & 66.76 (+10.73) & 63.22 (+7.20) & 75.80 (+19.78) \\
      & Duck
      & 72.07 & 78.63 (+6.56) & 78.07 (+6.00) & 76.09 (+4.02) & 83.65 (+11.58) \\
      \cmidrule(lr){2-7}
      & Overall
      & 59.77 & 69.75 (+9.98) & 71.69 (+11.93) & 67.97 (+8.20) & 80.21 (+20.44) \\

    \midrule
    \multirow{6}{*}{\makecell[c]{B100}}
      & Spray
      & 64.76 & 73.22 (+8.46) & 74.63 (+9.88) & 72.66 (+7.90) & 78.88 (+14.12) \\
      & Pringles
      & 20.14 & 40.93 (+20.79) & 44.88 (+24.74) & 33.33 (+13.19) & 68.98 (+48.83) \\
      & Tincase
      & 76.60 & 82.82 (+6.22) & 78.40 (+1.80) & 81.16 (+4.56) & 84.51 (+7.91) \\
      & Sandwich
      & 68.15 & 75.23 (+7.08) & 77.85 (+9.69) & 74.36 (+6.21) & 85.15 (+17.00) \\
      & Mouse
      & 59.98 & 69.90 (+9.93) & 68.37 (+8.39) & 64.10 (+4.12) & 76.17 (+16.20) \\
      & Duck
      & 71.95 & 79.28 (+7.33) & 76.98 (+5.02) & 76.21 (+4.26) & 83.60 (+11.65) \\
      \cmidrule(lr){2-7}
      & Overall
      & 60.26 & 70.23 (+9.97) & 70.18 (+9.92) & 66.97 (+6.71) & 79.55 (+19.29) \\
    \bottomrule
  \end{tabular}}
\end{table*}

\begin{table*}[h]
\centering
\caption{\textbf{AUC@[0:0.1] performance under diverse environments of various sensor control methods for SAM-6D.}}
\label{tab:unseen-environments-sam6d}
\resizebox{1.0\linewidth}{!}{
  \begin{tabular}{clccccc}
    \toprule
    \textbf{Env} & \textbf{Object} & \textbf{Baseline} & \textbf{Depth-Only} & \textbf{RGB-Only} & \textbf{Oracle-F} & \textbf{Oracle-D} \\
    \midrule

    \multirow{6}{*}{\makecell[c]{B5}}
      & Spray
      & 84.54 & 86.83 (+2.29) & 84.22 (-0.32) & 86.59 (+2.05) & 89.22 (+4.68) \\ 
      & Pringles
      & 50.17 & 71.45 (+21.28) & 69.14 (+18.97) & 65.76 (15.59) & 80.48 (+30.31) \\
      & Tincase
      & 47.61 & 66.73 (+19.12) & 77.96 (+30.34) & 68.61 (+21.00) & 80.67 (+33.05) \\
      & Sandwich
      & 78.36 & 79.00 (+0.64) & 80.10 (+1.74) & 78.41 (+0.05) & 80.85 (+2.49) \\
      & Mouse
      & 52.41 & 64.83 (+12.41) & 69.12 (+16.71) & 59.54 (+7.13) & 74.40 (+21.99) \\ 
      & Duck
      & 76.93 & 81.42 (+4.49) & 84.56 (+7.63) & 79.21 (+2.28) & 87.91 (10.98) \\
      \cmidrule(lr){2-7}
      & Overall
      & 65.00 & 75.04 (+10.04) & 77.52 (+12.51) & 73.02 (+8.02) & 82.25 (+17.25) \\

    \midrule
    \multirow{6}{*}{\makecell[c]{B25}}  
      & Spray
      & 82.00 & 85.90 (+3.90) & 77.80 (-4.20) & 82.68 (+0.68) & 89.49 (+7.49) \\
      & Pringles
      & 65.85 & 73.93 (+8.07) & 72.90 (7.05) & 66.95 (+1.10) & 80.67 (+14.81) \\
      & Tincase
      & 79.66 & 80.71 (+1.05) & 81.62 (+1.96) & 79.86 (+0.20) & 84.64 (+4.99) \\
      & Sandwich
      & 78.10 & 78.74 (+0.64) & 80.10 (+2.00) & 78.44 (+0.33) & 80.85 (+2.74) \\
      & Mouse
      & 55.90 & 62.59 (+6.69) & 66.90 (+11.00) & 62.50 (+6.60) & 72.32 (+16.42) \\
      & Duck
      & 75.65 & 82.40 (+6.74) & 82.74 (+7.09) & 79.07 (+3.42) & 87.47 (+11.81) \\
      \cmidrule(lr){2-7}
      & Overall
      & 72.86 & 77.38 (+4.52) & 77.01 (+4.15) & 74.92 (+2.06) & 82.57 (+9.71) \\

    \midrule
    \multirow{6}{*}{\makecell[c]{B50}}
      & Spray
      & 80.56 & 84.29 (+3.73) & 81.93 (+1.36) & 84.70 (+4.14) & 89.44 (+8.87) \\
      & Pringles
      & 67.45 & 74.74 (+7.29) & 75.95 (+8.50) & 71.26 (+3.81) & 81.31 (+13.86) \\
      & Tincase
      & 80.00 & 80.58 (+0.58) & 79.76 (-0.24) & 78.18 (-1.82) & 84.49 (+4.49) \\
      & Sandwich
      & 78.38 & 79.18 (+0.79) & 79.77 (+1.38) & 78.59 (+0.21) & 80.69 (+2.31) \\
      & Mouse
      & 56.35 & 62.76 (+6.41) & 66.90 (+10.55) & 59.80 (+3.45) & 71.68 (+15.33) \\
      & Duck
      & 77.21 & 82.60 (+5.40) & 83.23 (+6.02) & 80.02 (+2.81) & 87.70 (+10.49) \\
      \cmidrule(lr){2-7}
      & Overall
      & 73.33 & 77.36 (+4.03) & 77.92 (+4.60) & 75.43 (+2.10) & 82.55 (+9.22) \\

    \midrule
    \multirow{6}{*}{\makecell[c]{B75}}
      & Spray
      & 82.85 & 85.88 (+3.03) & 83.76 (+0.91) & 86.47 (+3.62) & 87.17 (+4.32) \\
      & Pringles
      & 67.48 & 75.38 (+7.90) & 76.62 (+9.14) & 70.45 (+2.98) & 80.29 (+12.81) \\   
      & Tincase
      & 81.14 & 80.69 (-0.45) & 82.87 (+1.73) & 81.18 (+0.05) & 84.18 (+3.04) \\  
      & Sandwich
      & 78.38 & 78.77 (+0.38) & 80.00 (+1.62) & 78.97 (+0.59) & 80.72 (+2.33) \\     
      & Mouse
      & 56.69 & 59.54 (+2.84) & 67.41 (+10.72) & 63.05 (+6.36) & 71.37 (+14.67) \\
      & Duck
      & 75.72 & 83.00 (+7.28) & 83.60 (+7.88) & 79.86 (+4.14) & 88.21 (+12.49) \\
      \cmidrule(lr){2-7}
      & Overall
      & 73.71 & 77.21 (+3.50) & 79.04 (+5.33) & 76.67 (+2.96) & 81.99 (+8.28) \\

    \midrule
    \multirow{6}{*}{\makecell[c]{B100}}
      & Spray
      & 85.67 & 85.20 (-0.48) & 85.49 (-0.19) & 86.12 (+0.45) & 87.20 (+1.52) \\
      & Pringles
      & 69.21 & 74.31 (+5.10) & 75.74 (+6.52) & 71.07 (+1.86) & 80.40 (+11.19) \\
      & Tincase
      & 78.79 & 78.93 (+0.14) & 78.73 (-0.06) & 80.77 (+1.98) & 82.60 (+3.81) \\
      & Sandwich
      & 78.74 & 79.28 (+0.54) & 80.13 (+1.38) & 79.55 (+0.81) & 80.92 (+2.18) \\
      & Mouse
      & 57.20 & 61.61 (+4.41) & 68.12 (+10.92) & 61.47 (+4.27) & 72.51 (+15.31) \\
      & Duck
      & 77.86 & 82.67 (+4.81) & 82.84 (+4.98) & 80.93 (+3.07) & 86.93 (+9.07) \\
      \cmidrule(lr){2-7}
      & Overall
      & 74.58 & 77.00 (+2.42) & 78.51 (+3.93) & 76.65 (+2.07) & 81.76 (+7.18) \\
    \bottomrule
  \end{tabular}}
\end{table*}

\begin{table*}[h]
\centering
\caption{\textbf{AUC@[0:0.1] performance under diverse environments of various sensor control methods for FoundationPose.}}
\label{tab:unseen-environments-foundationpose}
  \resizebox{1.0\linewidth}{!}{
  \begin{tabular}{clccccc}
    \toprule
    \textbf{Env} & \textbf{Object} & \textbf{Baseline} & \textbf{Depth-Only} & \textbf{RGB-Only} & \textbf{Oracle-F} & \textbf{Oracle-D} \\
    \midrule

    \multirow{6}{*}{\makecell[c]{B5}}
      & Spray
      & 82.80 & 84.68 (+1.88) & 85.34 (+2.54) & 83.44 (+0.64) & 87.17 (+4.37) \\
      & Pringles
      & 25.07 & 39.45 (+14.38) & 45.26 (+20.19) & 32.98 (+7.91) & 64.10 (+39.03) \\
      & Tincase
      & 26.47 & 49.11 (+22.64) & 53.64 (27.18) & 38.29 (+11.82) & 74.24 (+47.78) \\
      & Sandwich
      & 81.62 & 84.41 (+2.79) & 83.38 (+1.76) & 82.77 (+1.15) & 86.28 (+4.66) \\
      & Mouse
      & 67.88 & 72.80 (+4.92) & 75.10 (+7.22) & 72.02 (+4.14) & 79.54 (+11.66) \\
      & Duck
      & 82.58 & 85.21 (+2.63) & 87.88 (+5.30) & 84.07 (+1.49) & 89.74 (+7.16) \\
      \cmidrule(lr){2-7}
      & Overall
      & 61.07 & 69.28 (+8.21) & 71.77 (+10.70) & 65.59 (+4.52) & 80.18 (+19.11) \\
      
    \midrule
    \multirow{6}{*}{\makecell[c]{B25}}
      & Spray
      & 84.85 & 86.37 (+1.52) & 87.17 (+2.32) & 84.90 (+0.05) & 88.46 (+3.61) \\
      & Pringles
      & 40.24 & 51.57 (+11.33) & 51.36 (+11.12) & 44.67 (+4.43) & 71.07 (+30.83) \\
      & Tincase
      & 37.91 & 54.42 (+16.51) & 65.98 (+28.07) & 50.93 (+13.02) & 80.00 (+42.09)\\
      & Sandwich
      & 83.36 & 86.72 (+3.36) & 84.90 (+1.54) & 84.10 (+0.74) & 87.97 (+4.61) \\
      & Mouse
      & 71.39 & 75.37 (+3.98) & 75.98 (+4.59) & 72.66 (+1.27) & 80.51 (+9.12) \\
      & Duck
      & 84.74 & 87.16 (+2.42) & 88.65 (+3.91) & 85.63 (+0.88) & 90.40 (+5.65) \\
      \cmidrule(lr){2-7}
      & Overall
      & 67.08 & 73.60 (+6.52) & 75.67 (+8.59) & 70.48 (+3.40) & 83.07 (+15.99) \\
      
    \midrule
    \multirow{6}{*}{\makecell[c]{B50}}
      & Spray
      & 85.49 & 86.85 (+1.36) & 88.22 (+2.73) & 85.32 (-0.17) & 89.59 (+4.10) \\
      & Pringles
      & 35.67 & 55.31 (+19.64) & 50.74 (+15.07) & 45.50 (+9.83) & 79.00 (+43.33) \\
      & Tincase
      & 45.60 & 60.09 (+14.49) & 75.58 (+29.98) & 56.64 (+11.04) & 85.00 (+39.40) \\
      & Sandwich
      & 83.62 & 87.13 (+3.51) & 85.46 (+1.84) & 85.00 (+1.38) & 88.56 (+4.94) \\
      & Mouse
      & 70.71 & 74.39 (+3.68) & 75.78 (+5.07) & 72.51 (+1.80) & 79.88 (+9.17) \\
      & Duck
      & 85.23 & 87.33 (+2.09) & 90.19 (+4.95) & 85.74 (+0.51) & 91.42 (+6.19) \\     \cmidrule(lr){2-7}
      & Overall
      & 67.72 & 75.18 (+7.46) & 77.66 (+9.94) & 71.79 (+4.07) & 85.57 (+17.86) \\
      
    \midrule
    \multirow{6}{*}{\makecell[c]{B75}}
      & Spray
      & 85.93 & 87.20 (+1.27) & 88.73 (+2.80) & 86.12 (+0.19) & 89.83 (+3.90) \\
      & Pringles
      & 35.19 & 49.95 (+14.76) & 60.00 (+24.81) & 54.76 (+19.57) & 81.88 (+46.69) \\
      & Tincase
      & 41.87 & 58.62 (+16.76) & 75.73 (+33.87) & 55.31 (+13.44) & 87.27 (+45.40) \\
      & Sandwich
      & 83.56 & 86.82 (+3.26) & 85.82 (+2.26) & 84.92 (+1.36) & 88.92 (+5.36) \\
      & Mouse
      & 70.98 & 74.78 (+3.80) & 75.59 (+4.61) & 72.88 (+1.90) & 80.10 (+9.12) \\
      & Duck
      & 85.79 & 87.49 (+1.70) & 89.95 (+4.16) & 85.95 (+0.16) & 91.21 (+5.42) \\
      \cmidrule(lr){2-7}
      & Overall
      & 67.22 & 74.14 (+6.92) & 79.30 (+12.09) & 73.32 (+6.11) & 86.53 (+19.32) \\
      
    \midrule
    \multirow{6}{*}{\makecell[c]{B100}}
      & Spray
      & 84.07 & 87.51 (+3.44) & 88.54 (+4.47) & 86.00 (+1.93) & 90.00 (+5.93) \\
      & Pringles
      & 35.14 & 57.31 (+22.17) & 46.55 (+11.41) & 48.07 (+12.93) & 77.40 (+42.26) \\
      & Tincase
      & 39.36 & 58.56 (+19.20) & 75.96 (+36.60) & 55.09 (+15.73) & 87.42 (+48.07) \\
      & Sandwich
      & 83.69 & 86.92 (+3.23) & 86.03 (+2.34) & 85.21 (+1.52) & 88.82 (+5.13) \\
      & Mouse
      & 70.17 & 74.37 (+4.20) & 76.88 (+6.71) & 72.46 (+2.29) & 79.85 (+9.68) \\
      & Duck
      & 85.16 & 87.16 (+2.00) & 89.09 (+3.93) & 85.33 (+0.16) & 90.58 (+5.42) \\
      \cmidrule(lr){2-7}
      & Overall
      & 66.27 & 75.30 (+9.04) & 77.17 (+10.91) & 72.03 (+5.76) & 85.68 (+19.41) \\
    \bottomrule
  \end{tabular}}
\end{table*}

\clearpage
\section{ADD-VSD Correlation}
\label{appendix:add-vsd-corr}
\cref{tab:corr} reports correlation coefficients computed under the experimental settings described in~\cref{sec:unseen} and the PBR+Train-Def scheme of~\cref{sec:instance-level}. The correlations are first computed at the scene level across all sensor configurations, and then Fisher’s $z$-transformation is applied before averaging them across scenes.

\begin{table}[H]
\centering
\caption{\textbf{Correlation analysis between ADD and $AR_{\text{VSD}}$.}
$r$ denotes the Pearson correlation coefficient and $\rho_s$ denotes the Spearman rank correlation coefficient. For pretrained models, correlations are reported for both ADD(-S) and $AR_{\text{VSD}}$.}
\label{tab:corr}

\begin{subtable}[t]{\linewidth} 
\centering
\caption{Supervised Models}
\resizebox{\linewidth}{!}{
\begin{tabular}{cc @{\quad} cc @{\quad} cc}
\toprule
\multicolumn{2}{c}{\textbf{ZebraPose~\cite{zebrapose}}} 
& \multicolumn{2}{c}{\textbf{GDRNPP~\cite{gdrnpp}}} 
& \multicolumn{2}{c}{\textbf{HiPose~\cite{hipose}}} \\
\cmidrule(r){1-2} \cmidrule(r){3-4} \cmidrule(r){5-6}
$|r|$ & $|\rho_s|$ & $|r|$ & $|\rho_s|$ & $|r|$ & $|\rho_s|$ \\
\midrule
 0.867 & 0.947 & 0.935 & 0.923 & 0.982 & 0.938 \\
\bottomrule
\end{tabular}}
\end{subtable}
\begin{subtable}[t]{\linewidth} 
\centering
\caption{Pretrained Models.}
\resizebox{\linewidth}{!}{
\begin{tabular}{cc @{\quad} cc @{\quad} cc}
\toprule
\multicolumn{2}{c}{\textbf{GigaPose~\cite{gigapose}}} 
& \multicolumn{2}{c}{\textbf{SAM-6D~\cite{sam6d}}} 
& \multicolumn{2}{c}{\textbf{FondationPose~\cite{foundationpose}}} \\
\cmidrule(r){1-2} \cmidrule(r){3-4} \cmidrule(r){5-6}
$|r|$ & $|\rho_s|$ & $|r|$ & $|\rho_s|$ & $|r|$ & $|\rho_s|$ \\
\midrule
  0.846 & 0.805 & 0.718 & 0.715 & 0.696 & 0.621 \\
  (0.862) & (0.882) & (0.787) & (0.788) & (0.789) & (0.730) \\
\bottomrule
\end{tabular}}
\end{subtable}
\vspace{-1ex}
\end{table}

\newpage
\section{Detailed Result on HiPose}
\label{appendix:hipose_detail}
\cref{tab:hipose_detail} reports detailed object-wise results corresponding to \cref{tab:multimodal-comparion}.

\begin{table}[H]
\centering
\caption{\textbf{Object-wise AUC@[0:0.1] performance of various sensor control methods for HiPose, trained on PBR + Train-Def and tested on Test-Var.}}
\vspace{-2ex}
\label{tab:hipose_detail}
\resizebox{\linewidth}{!}{
\begin{tabular}{l c c c c c}
\toprule
\textbf{Object} & 
\makecell[c]{\textbf{Baseline}} &
\makecell[c]{\textbf{Depth-Only}} &
\makecell[c]{\textbf{RGB-Only}} &
\makecell[c]{\textbf{Oracle-F}} &
\makecell[c]{\textbf{Oracle-D}} \\
\midrule

Spray & 86.92 & 88.17 (+1.25) & 88.79 (+1.87) & 71.70 (-15.22) & 89.87 (+2.95) \\
Pringles &  73.40 & 76.18 (+2.78) & 79.08 (+5.68) & 64.23 (-9.16) & 81.55 (+8.15) \\
Tincase &  74.78 & 77.88 (+3.10) & 85.30 (+10.52) & 67.52 (-7.27) & 86.84 (+12.06) \\
Sandwich &  78.68 & 81.72 (+3.04) & 84.06 (+5.38) & 66.19 (-12.48) & 86.43 (+7.75) \\
Mouse & 63.92 & 66.41 (+2.49) & 72.19 (+8.27) & 53.25 (-10.67) & 75.21 (+11.29) \\
Duck & 70.27 & 72.94 (+2.67) & 86.67 (+16.40) & 65.54 (-4.73) & 88.46 (+18.19) \\
\midrule
Overall & 74.66 & 77.22 (+2.44) & 82.68 (+8.02) & 64.74 (-9.92) & 84.73 (+10.07) \\

\bottomrule
\end{tabular}}
\vspace{-2ex}
\end{table}

\end{document}